\documentclass[twoside,11pt]{article}

\usepackage{jmlr2e}

\usepackage{amsmath}
\usepackage{mathtools}

\usepackage{xspace}
\usepackage{graphicx}
\usepackage{subcaption}
\usepackage{wrapfig}

\usepackage{floatrow}

\usepackage[english]{babel}
\RequirePackage{algorithm}
\RequirePackage{algorithmic}

\newcommand{\R}{\mathbb{R}}

\newcommand{\Po}{\mathbb{P}}
\DeclareMathOperator*{\argmin}{arg\,min}

\newcommand{\E}{\mathbb{E}}

\newtheorem{property}[theorem]{\textbf{Proposition}}
 \newtheorem{assumption}[theorem]{\textbf{Assumption}}
\newtheorem{setting}[theorem]{\textbf{Setting}}

\DeclarePairedDelimiter\ceil{\lceil}{\rceil}

\jmlrheading{1}{2000}{1-48}{4/00}{10/00}{achdou00a}{Juliette Achdou, Joseph C. Lam, Alexandra Carpentier and Gilles Blanchard}

\ShortHeadings{A minimax near-optimal algorithm for adaptive rejection sampling}{Achdou et al.}
\firstpageno{1}

\begin{document}

\title{A minimax near-optimal algorithm for adaptive rejection sampling}

\author{\name Juliette Achdou \email juliette.achdou@gmail.com \\
       \addr Numberly (1000mercis Group)\\
       Paris, France
       \AND
       \name Joseph C. Lam \email joseph.lam@ovgu.de \\
       \addr Otto-von-Guericke University \\
  Magdeburg, Germany
  \AND
       \name Alexandra Carpentier \email alexandra.carpentier@ovgu.de \\
       \addr Otto-von-Guericke University \\
  Magdeburg, Germany
  \AND
       \name Gilles Blanchard \email gilles.blanchard@math.uni-potsdam.de \\
       \addr Potsdam University \\
  Potsdam, Germany
  }

\editor{}

\maketitle

\begin{abstract}%
  Rejection Sampling is a fundamental Monte-Carlo method. It is used to sample from distributions admitting a probability density function which can be evaluated exactly at any given point, albeit at a high
  computational cost. However, without proper tuning, this technique implies a high rejection rate. Several methods have been explored to cope with this problem, based on the principle of adaptively estimating the density by a simpler function, using the information of the previous samples. Most of them either rely on strong assumptions on the form of the density, or do not offer any theoretical performance guarantee. We give the first theoretical lower bound for the problem of adaptive rejection sampling and introduce a new algorithm which guarantees a near-optimal rejection rate in a minimax sense.
\end{abstract}



\section{Introduction}\label{intro}
The breadth of applications requiring independent sampling from a probability distribution is sizable. Numerous classical statistical results, and in particular those involved in machine learning, rely on the independence assumption.
For some densities, direct sampling may not be tractable, and the evaluation of the density at a given point may be costly. Rejection sampling (RS) is a well-known Monte-Carlo method for sampling from a density $f$ on $\mathbb{R}^d$ when direct sampling is not tractable (see~\citealp{von195113}, \citealp{devroye1986sample}). It assumes access to a density $g$, called  the proposal density, and a positive constant $M$, called the rejection constant, such that $f$ is upper-bounded by $Mg$, which is called the \textit{envelope}. Sampling from $g$ is assumed to be easy. At every step, the algorithm draws a proposal sample $X$ from the density $g$ and a point $U$ from the uniform distribution on $[0,1]$, and accepts $X$ if $U$ is smaller than the ratio of $f(X)$ and $Mg(X)$, otherwise it rejects $X$. The algorithm outputs all accepted samples, which can be proven to be independent and identically distributed samples from the density $f$. This is to be contrasted with Markov Chain Monte Carlo (MCMC) methods which produce a sequence of non dependent samples---and fulfill therefore a different objective. Besides, the application of rejection sampling includes variational inference: \cite{naesseth2016rejection,naesseth2017reparameterization} generalize the reparametrization trick to distributions which can be generated by rejection sampling.

\subsection{Adaptive rejection sampling}

\textit{Rejection sampling} has a very intuitive geometrical interpretation. Consider the variable $Z=(X,Mg(X)U)$, where $X$, $M$, $g$ and $U$ are defined as above. As shown in Figure \ref{fig:geom} in the Supplementary Material, $Z$ has a
uniform distribution on the region under the graph of $Mg$, and the sample is accepted if it falls into the region under the graph of $f$.
Conditional to acceptance, $Z$ is then drawn uniformly from the area under the graph of $f$. Thus $X$ is drawn from the distribution with density $f$.
The acceptance probability is the ratio of the two areas, $1/M$.
This means that the closer $g$ is to $f$ ---and $M$ to 1---, the more samples are accepted. The goal is hence to find a good envelope of $f$ in order to obtain a number of rejected samples as small as possible. In the absence of prior knowledge on the target density $f$, the proposal is
typically the uniform density on a set including the support of $f$ (here assumed to be compact), and the rejection constant $M$ is set as an upper bound on $f$. Consequently, this method leads to rejecting many samples in most cases and $f$ is evaluated many times uselessly.  



\textit{Adaptive rejection sampling} is a variant motivated by the high number of rejected samples mentioned above. Given $n$, a \textit{budget} of evaluations of $f$, the goal is to maximize $\hat{n}$, the number of output samples which have to be drawn independently from $f$. In other words, the ratio $\frac{n-\hat{n}}{n}$, also called \textit{rejection rate}, is to be made as small as possible, like in standard rejection sampling. To achieve this maximal number of output samples, adaptive rejection sampling methods gradually improve the proposal function and the rejection constant by using the information given by the evaluations of $f$ at the previous proposal samples. These samples are used to estimate---and tightly bound $f$ from above.

\subsection{Literature review}
\paragraph{Closely related works.}
A recent approach in \cite{erraqabi2016pliable}, pliable rejection sampling (PRS), allows sampling from multivariate densities satisfying mild regularity assumptions. In particular the function $f$ is of a given $s$-H\"older regularity. PRS is a two-step adaptive algorithm, based on the use of non-parametric kernel methods for estimating the target density. Assume that PRS is given a budget of $n$ evaluations of the function $f$. 
For a density $f$ defined on a compact domain, PRS first evaluates $f$ on a number $N<n$ of points uniformly drawn in the domain of $f$. It uses these evaluations to produce an estimate of the density $f$ using Kernel regression. Then it builds a proposal density using a high probability confidence bound on the estimate of $f$. The associated rejection constant is then the renormalization constant. The proposal density multiplied by the rejection constant is proven to be with high probability a correct envelope, i.e.,~an upper bound for $f$. PRS then applies rejection sampling $n-N$ times using such an envelope. 
This method provides with high probability a \textit{perfect sampler}, i.e.,~a sampler which outputs \textit{i.i.d.~samples from the density $f$}. It also comes with efficiency guarantees. Indeed in dimension $d$, if $s\leq 2$ and for $n$ large enough, PRS reaches an average rejection rate of the order of $\left({\log(nd)}/{n}\right)^{\frac{s}{3s+d}}$. This means that it asymptotically accepts almost all the samples. However, there is no guarantee that this rate might not be improved using another algorithm. Indeed, no lower bound on the rejection rate over all algorithms is presented.\\
Another recent related sampling method is A* sampling  \citep{maddison2014sampling}. It is close to the OS* algorithm from \cite{dymetman2012algorithm} and relies on an extension of the Gumbel-max trick. The trick enables the sampling from a categorical distribution over classes $i\in [1, \ldots, n]$ with probability proportional to $\exp(\phi(i))$, where $\phi$ is an unnormalized mass. It uses the following property of the Gumbel distribution. Adding Gumbel noise to each of the $\phi(i)$'s and taking the argmax of the resulting variables returns $i$ with a probability proportional to $\exp(\phi(i))$. Then, the authors generalize the notion of Gumbel-max trick to a continuous distribution. 
This method shows good empirical efficiency in the number of evaluations of the target density. However, the assumption that the density can be decomposed into a bounded function and a function, that is easy to integrate and sample from, is rarely true in practice.

\paragraph{Other related works.}
\cite{gilks1992adaptive} introduced ARS: a technique of adaptive rejection sampling for one-dimensional log-concave and differentiable densities whose derivative can be evaluated.
ARS sequentially builds a tight envelope of the density by exploiting the concavity of $\log(f)$ in order to bound it from above. At each step, it samples a point from a proposal density. It evaluates $f$ at this point, and updates the current envelope to a new one which is closer to $f$. The proposal density and the envelope thus converge towards $f$, while the rejection constant converges towards 1. The rejection rate is thereby improved. 
 \cite{gilk1992derivative} also developed an alternative to this ARS algorithm for the case where the density is not differentiable or the derivative can not be evaluated. The main difference with the former method is that the computation of the new proposal does not require any evaluation of the derivative. For this algorithm, as for the one presented in \cite{gilks1995adaptive}, the assumption that the density is log-concave represents a substantial constraint in practice. In particular, it restrains the use of ARS to unimodal densities. \\
An extension from \cite{hormann1995rejection} of ARS adapts it to $T$-concave densities, with $T$ being a monotonically increasing transformation. However, this method still cannot be used with multimodal densities. 
In 1998, Evans and Swarz proposed a method applicable to multimodal densities presented in \cite{evans1998random} which extends the former one. It deals with $T$-transformed densities and spots the intervals where the transformed density is concave or convex. Then it applies an ARS-like method separately on each of these intervals. However it needs access to the inflection points, which is a strong requirement. A more general method in \cite{gorur2011concave} consists of decomposing the log of the target density into a sum of a concave and convex functions. It deals with these two components separately. An obvious drawback of this technique is the necessity of the decomposition itself, which may be a difficult task.
Similarly, \cite{martino2011generalization} deal with cases where the log-density can be expressed as a sum of composition of convex functions and of functions that are either convex or concave. This represents a relatively broad class of functions; other variants focusing on the computational cost of ARS have been explored in \cite{martino2017parsimonious,martino2017adaptive}.\\
For all the methods previously introduced, no theoretical efficiency guarantees are available.


A further attempt at improving simple rejection sampling resulted in Adaptive Rejection Metropolis Sampling (ARMS) \citep{gilks1995adaptive}. ARMS extends ARS to cases where densities are no longer assumed to be log-concave. It builds a proposal function whose formula is close to the one in \cite{gilk1992derivative}. This time however, the proposal might not be an envelope, which would normally lead to oversampling in the regions where the proposal is smaller than the density. In ARMS, this is compensated with a Metropolis-Hastings control-step. 
One drawback of this method is that it outputs a Markov Chain, in which the samples are correlated. Moreover, the chain may be trapped in a single mode. 
Improved adaptive rejection Metropolis (\citealp{martino2012improved}) modifies ARMS in order to ensure that the proposal density tends to the target density. In \cite{meyer2008adaptive} an alternative is presented that uses  polynomial interpolations as proposal functions. However, this method still yields correlated samples. 

Markov Chain Monte Carlo (MCMC) methods (\citealp{metropolis1949monte, andrieu2003introduction}) represent a very popular set of generic approaches in order to sample from a distribution. Although they scale with dimension better than rejection sampling, they are not perfect samplers, as they do not produce i.i.d.~samples, and can therefore not be applied to achieve our goals. Variants producing independent samples were proposed in \cite{fill1997interruptible,propp1998coupling}. However, to the best of our knowledge, no theoretical studies on the rejection rate of these variants is available in the literature.

Importance sampling is a problem close to rejection sampling, and adaptive importance sampling algorithms are also available (see e.g.,~\citealp{oh1992adaptive,cappe2008adaptive,ryu2014adaptive}). Among them, the algorithm in \cite{zhang1996nonparametric} sequentially estimates the target function, whose integral has to be computed using kernel regression, similarly to the approach of \cite{erraqabi2016pliable}. A recent notable method regarding discrete importance sampling was introduced in \cite{canevet2016importance}. In \cite{delyon2018efficiency}, adaptive importance sampling is shown to be efficient in terms of asymptotic variance. 

\subsection{Our contributions}

The above mentioned sampling methods either do not provide i.i.d samples, or do not come with theoretical efficiency guarantees, apart from \cite{erraqabi2016pliable} or \cite{zhang1996nonparametric, delyon2018efficiency} in importance sampling. In the present work, we propose the Nearest Neighbour Adaptive  Rejection Sampling algorithm (NNARS), an adaptive rejection sampling technique which requires $f$ to have $s$-H\"older regularity (see Assumption~\ref{ass:f}). Our contributions are threefold, since NNARS:
\begin{itemize}
\item is a \textit{perfect sampler} for sampling from the density $f$. 
\item offers an \textit{average rejection rate of order $\log(n)^2n^{s/d}$}, if $s\leq 1$. This significantly improves the state of the art average rejection rate from~\cite{erraqabi2016pliable} over $s$-H\"older densities, which is of order $\left({\log(nd)}/{n}\right)^{\frac{s}{3s+d}}$.
\item matches a \textit{lower bound for the rejection rate} on the class of all adaptive rejection sampling algorithms and all $s$-H\"older densities. It gives an answer to the theoretical problem of quantifying the difficulty of adaptive rejection sampling in the minimax sense. So NNARS offers a near-optimal average rejection rate, in the minimax sense over the class of H\"older densities.
\end{itemize}
NNARS follows a common approach to that of most adaptive rejection sampling methods. It relies on non-parametric estimation of $f$. It improves this estimation iteratively, and as the latter gets closer to $f$, the envelope also approaches $f$. Our improvements consist of designing an optimal envelope, and updating the envelope as we get more information at carefully chosen times. This leads to an average rejection rate for NNARS which is minimax near-optimal (up to a logarithmic term) over the class of H\"older densities. No adaptive rejection algorithm can perform significantly better on this class. The proof of the minimax lower bound is also new to the best of our knowledge.

The optimal envelope we construct is a very simple one. For every known point of the target density $f$, we use the regularity assumptions on $f$ in order to construct an envelope which is piecewise constant. It stays constant in the neighborhood of every known point of $f$. Figure~\ref{fig:NNARS} 
depicts NNARS' first steps on a mixture of Gaussians in dimension 1.

In the second section of this paper, we set the problem formally and discuss the assumptions that we make. In the third section, we introduce the NNARS algorithm and provide a
theoretical upper bound on its rejection rate. In the fourth section, we present a minimax lower bound for the problem of adaptive rejection sampling. In the fifth section, we discuss our method and detail the open questions regarding NNARS. In the sixth section, we present experimental results on both simulated and real data that compare our strategy with state of the art algorithms for adaptive rejection sampling. Finally, the Supplementary Material contains the proofs of all the results presented in this paper.

\section{Setting}
Let $f$ be a bounded density defined on $[0,1]^d$. The objective is to provide an algorithm which outputs as many i.i.d.~samples drawn according to $f$ as possible, with a fixed number $n$ of evaluations of $f$. We call $n$ the budget.

\subsection{Description of the problem}

The framework that we consider is \textit{sequential and adaptive rejection sampling}.
\paragraph{Adaptive Rejection Sampling (ARS).} Set $\mathcal S = \emptyset$ and let $n$ be the budget. An ARS method sequencially performs $n$ steps 
At each step $t \leq n$, the samples $\{X_1,\ldots,X_{t-1}\}$ collected until $t$, each in $[0,1]^d$, are known to the learner, as well as their images by $f$.
 The learner $A$ chooses a positive constant $M_t$ and a density $g_t$ defined on $[0,1]^d$ that both depend on the previous samples and on the evaluations of $f$ at these points $\left\{(X_1, f(X_1)),\ldots, (X_{t-1}, f(X_{t-1}))\right\}$. Then the learner $A$ performs a rejection sampling step with the proposal and rejection constant $(g_t, M_t)$, as depicted in Algorithm~\ref{algo:RS}. It generates a point $X_t$ from $g$ and a variable $U_t$ that is independent from every other variable and drawn uniformly from $[0,1]$. $X_t$ is accepted as a sample from $f$ if $U_t\leq \frac{f(X_t)}{M_t g_t(X_t)}$ and rejected otherwise. If it is accepted, the output is $X_t$, otherwise the output is $\emptyset$.
Once the rejection sampling step is complete, the learner adds the output of this rejection sampling step to $\mathcal{S}$.
The learner iterates until the budget $n$ of evaluations of $f$ has been spent.
\begin{algorithm}
\caption{Rejection Sampling Step with ($f,g,M$) : RSS($f,g,M$)}
\label{algo:RS}
\begin{algorithmic}
 \STATE {\bfseries Input:} target density $f$,  proposal density $g$, rejection constant $M$.
 \STATE {\bfseries Output:} Either a sample $X$ from $f$, or nothing.
 \STATE Sample $X \sim g$ and $U \sim \mathcal{U}_{[0,1]}$.
 \IF{$ U \leq \frac{f(X)}{M g(X)}$} \STATE output $X$.
 \ELSE
 \STATE output $\emptyset$.
 \ENDIF
\end{algorithmic}
\end{algorithm}

\noindent
\begin{minipage}{\textwidth}
\begin{definition}\label{game:1} {\bf (Class of Adaptive Rejection Sampling (ARS) Algorithms)}~\\
An algorithm $A$ is an ARS algorithm if, given $f$ and $n$, at each step t $\in \{1\ldots n\}$:
\begin{itemize}
\item $A$ chooses a density $g_t$, and a positive constant $M_t$, depending on:\\ $\Big\{(X_1, f(X_1)), \ldots, (X_{t-1},f(X_{t-1}))\Big\}$.
\item $A$ performs a Rejection Sampling Step with ($f, g_t,M_t$).
\end{itemize}
The objective of an ARS algorithm is to sample as many i.i.d.~points according to $f$ as possible.
\end{definition}
\end{minipage}

\begin{theorem}\label{th:ARSa}
Given access to a positive, bounded density $f$ defined on $[0,1]^d$, any Adaptive Rejection Sampling algorithm (as described above) satisfies:\\
if $~ \forall t \leq n, ~ \forall x\in[0,1]^d, ~ f(x) \leq M_t g_t(x),$ the output
$\mathcal{S}$ contains i.i.d.~samples drawn according to $f$.
\end{theorem}

\paragraph{Definition of the loss.}
Theorem~\ref{th:ARSa} gives a necessary and sufficient condition under which an adaptive rejection sampling algorithm is a \textit{perfect sampler}---that is, it outputs i.i.d.~samples. Its proof is given in the Supplementary Material, see Appendix~\ref{proof:ARSa}.

Let us define the number of samples which are known to be independent and sampled according to $f$ based on Theorem \ref{th:ARSa}: $\hat{n} = \#\mathcal S\times \mathbf 1\{\forall t\leq n:\; f \leq M_t g_t\}$. We define the loss of the learner as $L_n=  n-\hat{n}$. This is justified by considering two complementary events. In the first, the rejection sampling procedure is correct at all steps, that is to say all proposed envelopes bound $f$ from above; and in the second, there exists a step where the procedure is not correct. In the first case, the sampler will output i.i.d.~samples drawn from the density $f$. So the loss of the learner $L_n$ is the number of samples rejected by the sampler.
In the second case, the sampler is not trusted to produce correct samples. So the loss becomes $n$,.
Finally, we note that the rejection rate is ${L_n}/{n}$.
\paragraph{Remark on the loss.}
Let $\mathcal{A}$ be the set of ARS algorithms defined in Definition~\ref{game:1}. Note that for any algorithm $A\in \mathcal A$, the loss $L_n(A)$ can be interpreted as a \textit{regret}. Indeed, a learner that can sample directly from $f$ would not reject a single sample, and would hence achieve
$L_n^* = 0.$
So $L_n(A)$ is equal to the difference between $L_n(A)$ and $L_n^*$. Hence $L_n(A)$ is the cost of not knowing how to sample directly from $f$.

\subsection{Assumptions}
We make the following assumptions on $f$. They will be used by the algorithm and for the theoretical results.

\begin{assumption} $\;$\vspace{1 mm}
\label{ass:f}
\begin{itemize}
\item The function $f$ is $(s,H)$-H\"older for some $0<s\leq 1$ and $H\geq 0$, \\
i.e., $ \forall x,y \in [0,1]^d , \; |f(x)-f(y)| \leq H \|x-y\|_{\infty}^s,$
where $\|u\|_{\infty} = \max_i |u_i|$;
\item There exists $0<c_f\leq 1$ such that: $\forall x \in [0,1]^d,\; c_f<f(x).$
\end{itemize}
\end{assumption}

Let $\mathcal{F}_0:= \mathcal{F}_0(s, H, c_f,d)$
denote the set of functions satisfying Assumption \ref{ass:f} for given $0<s\leq 1$, $H\geq 0$ and $0<c_f\leq 1$.\\ \
\paragraph{Remarks.} Here the domain of $f$ is assumed to be $[0,1]^d$, but it could without loss of generality be relaxed to any hyperrectangle of $\R^d$. Besides, for any distribution with sub-Gaussian tails, this assumption is almost true. In practice, the diameter of the support is bounded by $O(\sqrt{\log n})$, where $n$ is the number of evaluations, because of the vanishing tail property. The assumption of H\"older regularity is a usual regularity assumption in order to control for rates of convergence. It is also a mild one, considering that $s$ can be chosen arbitrarily close to $0$. Note however that we assume the knowledge of $s$ and $H$ for the NNARS algorithm.
 Since $f$ is a H\"older regular density defined on the unit cube, we can obtain the following upper bound: $f(x)\leq 1 + H \; \forall x \in [0,1]^d$. As for the assumption involving the constant $c_f$, it is widespread in non-parametric density estimation. Besides, the algorithm will still produce exact independent samples from the target distribution without the latter assumption.
 It is important to note that $f$ is chosen as a probability density for clarity, but it is not a required hypothesis. In the proofs, we study the general case when $f$ is not assumed to be a probability density.

\section{The NNARS Algorithm}\label{sec:algo}

The NNARS algorithm proceeds by constructing successive proposal functions $g_t$ and rejection constants $M_t$ that gradually approach $f$ and $1$, respectively. In turn, an increased number of evaluations of $f$ should result in a more accurate estimate and thus in a better upper bound.

\subsection{Description of the algorithm}

 The algorithm outlined in Algorithm~\ref{algo:NNARS} takes as inputs the budget $n$, and $c_f, H, s$ as defined in Assumption \ref{ass:f}. Let $\mathcal S$ denote its output. At each round $k$, the algorithm performs a number of RSS steps with specifically chosen $g_k$ and $M_k$. We call $\chi_k$ the set of points generated at round $k$ and of their images by $f$, whether they get accepted or rejected.\\ 
\paragraph{Initialization.}~The sets $\mathcal S$ and $\chi_k, k \in \mathbb N$ are initialized to $\emptyset$. $g_1$ is a uniform proposal on $[0,1]^d$. $M_1=1+H$ is an upper bound on $f$ and $N = N_1 =  \lceil 2(10 H)^{d/s} \log(n) c_f^{-1-d/s} \rceil$.
For any function $h$ defined on $[0,1]^d$, we set $I_h = \int_{[0,1]^d} h(x) dx$.
\paragraph{Loop.}
The algorithm proceeds in 
$K = \left \lceil \log_p(\frac{n}{N}) \right \rceil$ rounds, where
$p= \left\lceil \frac{3}{2} \frac{1}{c_f} \right\rceil$,
$\lceil \; \rceil$ is the ceiling function, and $\log_p$ is the logarithm in base $p$.\\
Each round $k \in \{1, \ldots, K\}$ consists of the following steps.

\begin{enumerate}
    \item Perform a Rejection Sampling Step RSS($f,{g}_{k},{M}_{k}$)
  $N_k$ times.
  Add the accepted samples to $\mathcal{S}$.
  All proposal samples as well as their images by $f$ produced in the Rejection Sampling Step are stored in $\chi_k$, whether they are rejected or not.
  \item Build an estimate $\hat f_{\cup_{i \leq k} \chi_i}$ of $f$ based on the evaluations of $f$ at all points stored in $\cup_{i\leq k} \chi_i$, thanks to the Approximate Nearest Neighbor Estimator, referred to in Definition~\ref{def:estimator}, applied to the set $\chi_k$. 
  \item Compute the proposal with the formula:
\begin{align}\label{eq:gk}
{g}_{k+1}(x) = \frac{\hat f_{\cup_{i \leq k} \chi_i}(x) + \hat r_{\cup_{i \leq k} \chi_i}}{I_{\hat f_{\cup_{i \leq k} \chi_i}} + \hat r_{\cup_{i \leq k} \chi_i}},
\end{align}
and the rejection constant with the formula: 
\begin{align}\label{eq:Mk}
{M}_{k+1} = I_{\hat f_{\cup_{i \leq k} \chi_i}} + \hat r_{\cup_{i \leq k} \chi_i},
\end{align}
where $\hat r_{\cup_{i \leq k} \chi_i}$ is defined in Equation~\eqref{eq:rate} below, in Definition~\ref{def:estimator}. Note that $g_{k+1}$ and $M_{k+1}$ are indexed here by the number of the round, unlike in the last section where the index was the current time.
\item If $k<K-1$, set $N_{k+1}$ as $p N_{k} = p^{k} N.$ Otherwise
  $N_K = n - \frac{1-p^{K-1}}{1-p} N$.
\end{enumerate}
Finally, the algorithm outputs $\mathcal{S}$, the set of accepted samples that have been collected. 
 
\begin{definition}[Approximate Nearest Neighbor Estimator applied to $\chi$]\label{def:estimator} $\;$\\
Let $f$ be a positive density satisfying Assumption \ref{ass:f}. We consider a set of $\tilde N$ points and their images by $f$, $\chi =\{(X_1,f(X_1)),\ldots, (X_{\tilde N}, f(X_{\tilde N})))\}$. Let us define a set of centers of cells constituting a uniform grid of $[0,1]^d$, namely
$$\mathcal C_{\tilde N} = \left\{2^{-1}(\lfloor \tilde N^\frac{1}{d} \rfloor +1)^{-1}u, u \in \{1, \ldots, 2(\lfloor \tilde N^\frac{1}{d} \rfloor +1) - 1\}^d\right\}.$$
The cells are of side-length $1/(\lfloor N^\frac{1}{d} \rfloor +1)$. For $x\in [0,1]^d$,  write $ C_{\tilde N}(x) = \argmin_{u \in \mathcal C_{\tilde N}} \|x - u\|_\infty$.

  
We define the approximate nearest neighbor estimator as the piecewise-constant estimator $\hat{f}_{\chi}$ of f by:
$\forall x \in [0,1]^d, \; \hat{f}_{\chi}(x) = \hat{f}_{\chi}(C_{\tilde N}(x)) = f \left( X_{i(C_{\tilde N}(x))} \right) ,$
where $i(x) =  \displaystyle{\argmin_{i\leq \tilde N}(\|x-X_i\|_{\infty})} $.

We also write
\begin{align}\label{eq:rate}
\hat{r}_{\chi} = H \left(\max_{u \in \mathcal C_{\tilde N}} \min_{i \leq \tilde N} \|u - X_i\|_\infty + \frac{1}{2 (\lfloor \tilde N^\frac{1}{d} \rfloor +1)} \right)^s.
\end{align} 
\end{definition}
\paragraph{Remarks on the proposal densities and rejection constants}~ At each step, the envelope is made up of evaluations of $f$ summed with a positive constant which stands for a confidence term of the estimation. It provides an upper bound for $f$. Furthermore, the use of nearest neighbour estimation in a noiseless setting implies that this bound is optimal. Besides, the approximate construction of the estimator builds proposal densities which are simple to sample from.\\
\indent As explained in Lemma~\ref{lem:event} in the Supplementary Material, an important remark is that the proposal density $g_k$ from Equation~\eqref{eq:gk} multiplied by the rejection constant $M_k$ from Equation~\eqref{eq:Mk} is an envelope of $f$. This means $M_kg_k \geq f$ for all $k\leq K$. So by Theorem~\ref{th:ARSa}, NNARS is a perfect sampler.

\begin{figure}[H]
\includegraphics[width=0.45\textwidth]{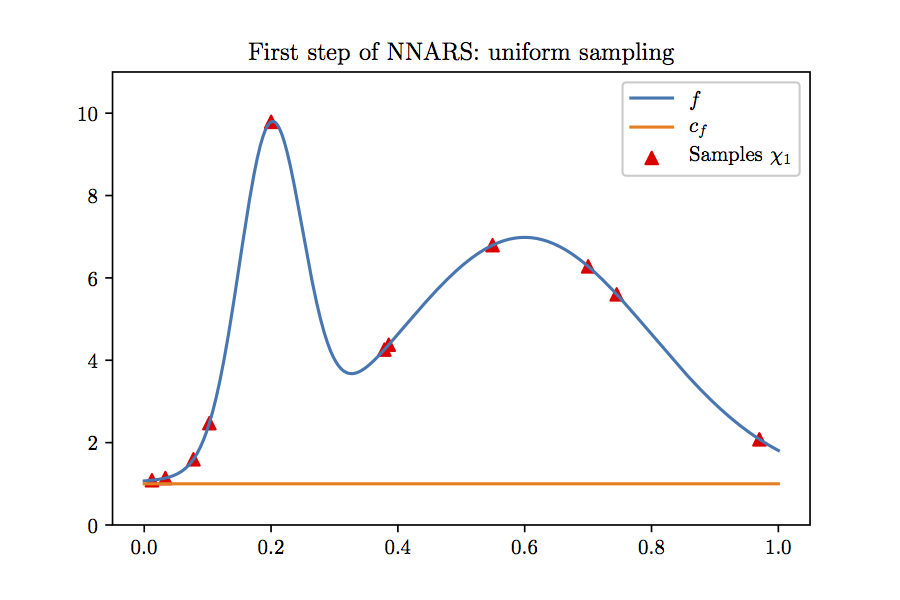}
	\includegraphics[width=0.45\textwidth]{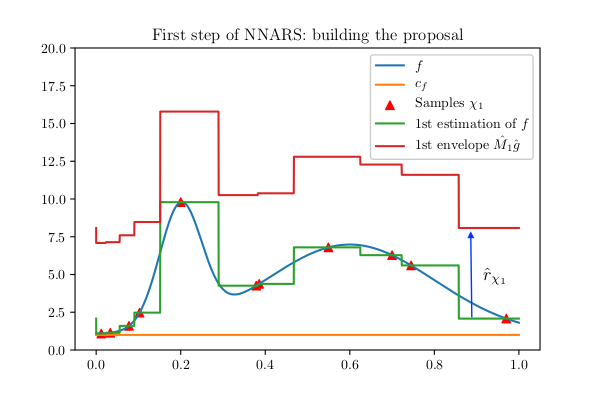}
	\includegraphics[width=0.45\textwidth]{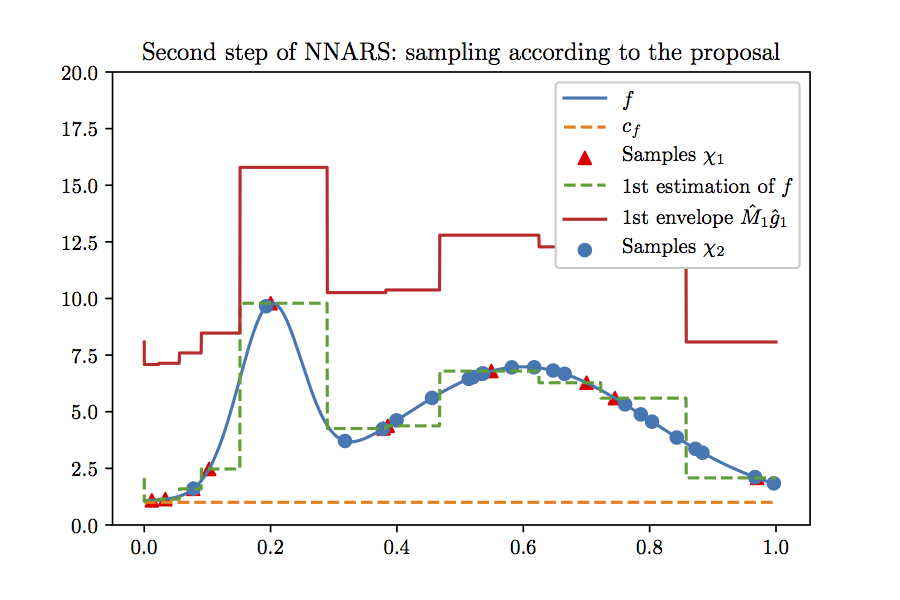}
    \includegraphics[width=0.45\textwidth]{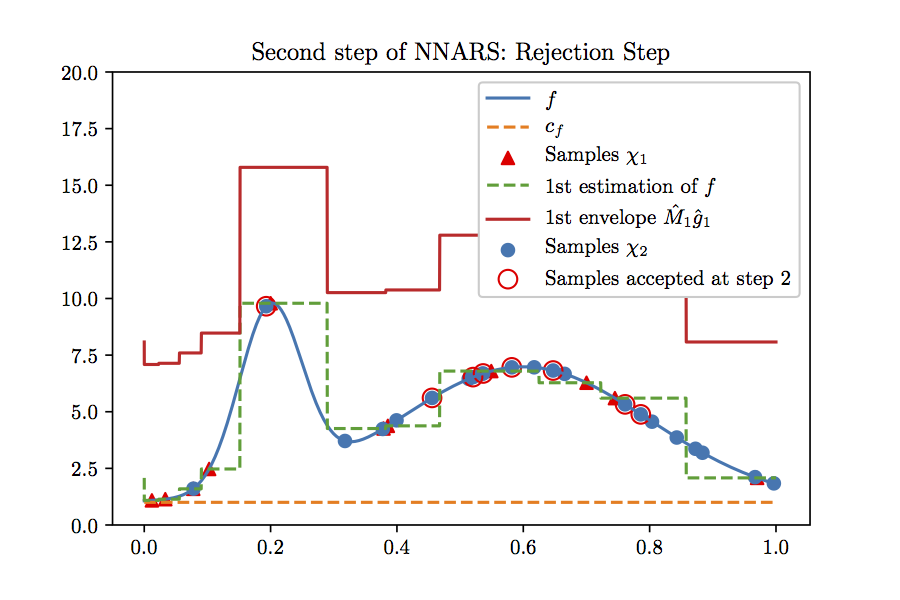}
    	\includegraphics[width=0.45\textwidth]{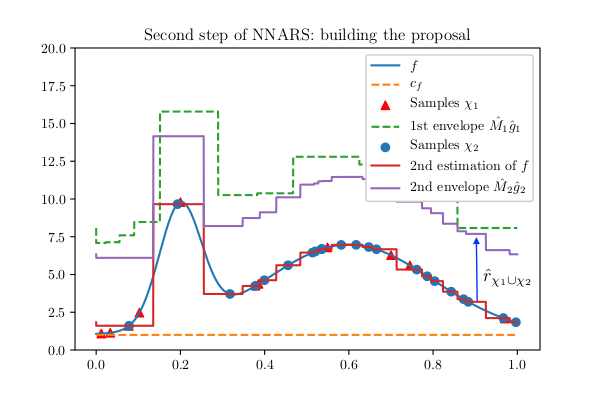}
	\caption{NNARS' first steps on a mixture of Gaussians (to be read in natural reading direction)}
	\label{fig:NNARS}
\end{figure}

The algorithm is illustrated in Figure \ref{fig:NNARS}.

\begin{algorithm}
\caption{Nearest Neighbor Adaptive Rejection Sampling}   
\label{algo:NNARS}
\begin{algorithmic}
  \STATE {\bfseries Input:} the budget $n$; the constants $H$, $s$ and $c_f$; the dimension $d$.
  \STATE {\bfseries Output:} the set $\mathcal{S}$ of i.i.d.~samples from $f$.
  \STATE Initialize $\mathcal{S} = \emptyset$,  $\chi_k = \emptyset \; \forall k$.
  \STATE Set $N_1=N$, ${g}_1 = \mathcal{U}_{[0,1]^d}$, $M_1 = 1+ H$.
\FOR {$k=1$ {\bfseries to} $K$}
  \FOR {$i=1$ {\bfseries to} $N_k$}
    \STATE Perform a Rejection Sampling Step RSS($f, {g}_k, {M}_{k}$).
    \STATE Add the output of RSS to $\mathcal S$.
    \STATE Add to $\chi_k$ both the sample from ${g}_{k}$ collected in the RSS, and its image by $f$.
  \ENDFOR
  \STATE Estimate $\hat f_{\cup_{i \leq k} \chi_k}$ according to Definition~\ref{def:estimator}.
  \STATE Compute ${g}_{k+1}$ and ${M}_{k+1}$ as in Equations~\eqref{eq:gk} and~\eqref{eq:Mk}.
\ENDFOR
\end{algorithmic}
\end{algorithm}

\noindent
{\bf Remark on sampling from the proposal densities $g_k$ in NNARS.} The number of rounds is of order $\lfloor \log(n) \rfloor$. The construction of the proposal in NNARS involves at each round $k$ the storage of $|\cup_{i\leq k} \chi_i| \propto p^{k+1} \lfloor \log(n) \rfloor$ values. So the total number of values stored is upper bounded by the budget. At each round, each value corresponds to a hypercube of side-length $1/|\cup_{i \leq k} \chi_i|^{1/d}$ splitting $[0,1]^d$ equally. Partitioning the space in this way allows us to efficiently assign a value to every $x \in [0,1]^d$, depending on which cell of the grid $x$ belongs to. Besides, sampling from the proposal amounts to sampling from a multinomial convolved with a uniform distribution on a hypercube. In other words, a cell is chosen multinomially, then a point is sampled uniformly inside that cell, because the proposal is piecewise constant.

The process to sample according to $g_{k}$ is the following: given $\cup_{i \leq k} \chi_i$, 
\begin{enumerate}
\item Each center of the cells from the grid $u \in \mathcal C_{|\cup_{i \leq k} \chi_i|}$ is mapped to a value $g_k(u)$.
\item One of the centers $\tilde C \in \mathcal C_{|\cup_{i \leq k} \chi_i|}$ is sampled with probability $g_k(\tilde C)$.
\item The sample point is drawn according to the uniform distribution on the hypercube of center $\tilde C$ and side-length $1/|\cup_{i \leq k} \chi_i|^{1/d}$.
\end{enumerate}


\subsection{Upper bound on the loss of NNARS}

In this section, we present an upper bound for the expectation of the loss of the NNARS sampler. This bound holds under Assumption \ref{ass:f}, that only requires $n$ to be large enough in comparison with constants depending on $d$, $s$, $c_f$ and $H$. Related conditions about the sample size are in most theoretical works on Rejection Sampling (\citealp{gilks1992adaptive}, \citealp{meyer2008adaptive}, \citealp{gorur2011concave}, \citealp{erraqabi2016pliable}).
\begin{assumption}[Assumption on $n$]\label{ass:n} $\;$\\
Assume that $n \geq 8$ and $N/n \leq 1/(2K^2)$, i.e.,
$$n \geq \left\lceil 2(10 H)^{d/s} \log(n) c_f^{-1-d/s} \right\rceil \frac{4\log(n)^2}{ \log\left(\left\lceil \frac{3}{2} \frac{1}{c_f} \right\rceil \right)^2} = O(\log(n)^3) .$$
\end{assumption}

\begin{theorem}\label{th:speedrate}
  Let $0<s\leq 1$, $H\geq 0$ and $c_f>0$. If $f$ satisfies Assumption \ref{ass:f} with $(s,H,c_f)$ such that $f\in \mathcal F_0(s,H,c_f,d)$, then NNARS is a perfect sampler according to $f$. 
  
 Besides if $n$ satisfies Assumption \ref{ass:n}, then 
\begin{align*}
 \E_f   L_n(\text{NNARS})  &\leq 
 40Hc_f^{-1}(1+\sqrt{2\log(3n)})(\log(2n))^{s/d}n^{1-s/d}
 \\ + 
& ~~~   \left(25 + 80c_f^{-1} + 2(10H)^{d/s}c_f^{-1-d/s}\right) \log^2(n)  = O(\log^2(n) n^{1-s/d}),
\end{align*}

\noindent
where $\E_f   L_n(\text{NNARS})$ is the expected loss of NNARS on the problem defined by $f$.
The expectation is taken over the randomness of the algorithm. This result is uniform over $\mathcal F_0(s,H,c_f,d)$.
\end{theorem}
The proof of this theorem is in the Supplementary Material, see Section~\ref{proof:speedrate}. The loss presented here divided by $n$ is to be interpreted as an upper bound for the expected rejection rate obtained by the NNARS algorithm. 



\paragraph{Sketch of the proof.}
The average number of rejected samples is $\sum_k N_k(1-1/M_k)$, since a sample is accepted at round $k$ with probability $1-1/M_k$. In order to bound the average number of rejected samples, we bound $M_k$ at each round $k$ with high probability.

By H\"older regularity and the definition of $\hat r_{\cup_{i \leq k} \chi_i}$ in Equation~\eqref{eq:rate} (in Definition~\ref{def:estimator}), we always have $|\hat f_{\cup_{i \leq k} \chi_i} -f | \leq \hat r_{\cup_{i \leq k} \chi_i}$, as shown in the proof of Proposition~\ref{pr:estimator}. So $M_k= I_{\hat f_{\cup_{i \leq k-1} \chi_i}} + \hat r_{\cup_{i \leq k-1} \chi_i}\leq I_f+2\hat r_{\cup_{i \leq k-1} \chi_i}$ with $I_f=1$.
Then, we consider the event $\mathcal{A}_{k,\delta} = \{ \forall j\leq k, \; \hat r_{\cup_{i \leq j}}  \leq C_0 H (\log(N_j/\delta)/N_j)^{s/d}\}$, where $C_0$ is a constant. Now, for each $k$, on $\mathcal{A}_{k-1,\delta}$, $M_k$ is bounded from above, with a bound of the order of $(\log(N_{k-1}/\delta)/N_{k-1})^{s/d}$. So, on $\mathcal{A}_{K,\delta}$, the average number of rejected samples has an upper bound of the order of $\log(n)^{2} n^{1-s/d}$, as presented in Theorem~\ref{th:speedrate}.

Now, we prove by induction that event $\mathcal{A}_{k,\delta}$ has high probability, as in the proof of Lemma~\ref{lem:bounded_g}. More precisely, $\mathcal{A}_{k,\delta}$ has probability larger than $1-2k\delta$. At every step $k$, we verify that $g_k$ is positively lower bounded conditionally on $\mathcal{A}_{k-1,\delta}$. Hence, the probability of having drawn at least one point in each hypercube of the grid with centers $\mathcal C_{|\cup_{i \leq k} \chi_i|}$ is high, as shown in the proof of Proposition~\ref{pr:rhat_r}. So the distance from any center to its closest drawn point is upper bounded with high probability. And this implies that $\mathcal{A}_{k,\delta}$ has high probability if $\mathcal{A}_{k-1,\delta}$ has high probability, which gets the induction going. On the other hand, the number of rejected samples is always bounded by $n$ on the small probability event where $\mathcal{A}_{K,\delta}$ does not hold. This concludes the proof.





\endproof
\section{Minimax Lower Bound on the Rejection Rate}
\label{sec:lower_bound}

It is now essential to get an idea of how much it is possible to reduce the loss obtained in Theorem~\ref{th:speedrate}. That is why we apply the framework of minimax optimality and complement the upper bound with a lower bound.
The minimax lower bound on this problem is the infimum of the supremum of the loss of algorithm $A$ on the problem defined by $f$; the infimum is taken over all adaptive rejection sampling algorithms $A$ and the supremum over all functions $f$ satisfying Assumption \ref{ass:f}. It characterizes the difficulty of the rejection sampling problem. And it provides the best rejection rate that can possibly be achieved by such an algorithm in a
worst-case sense over the class $\mathcal{F}_0(s,H,c_f,d)$.
\begin{theorem}\label{th:lower_bound}
For $0<s\leq 1$, there exists a constant $N(s,d)$ that depends only on $s,d$ and such that for any $n \geq N(s,d)$:
\begin{equation*}
\inf_{A \in \mathcal A}\; \sup_{f  \in \mathcal{F}_0(s,1,1,d)\cap \{f : I_f =1\}} \E_f (L_n(A)) \geq
   3^{-1}2^{-1-3s-2d}5^{-s/d}n^{1-s/d} = O(n^{1-s/d}),    
\end{equation*}
where $\E_f(L_n(A))$ is the expectation of the loss of $A$ on the problem defined by $f$. It is taken over the randomness of the algorithm $A$.
\end{theorem}

\noindent
The proof of this theorem is in Section~\ref{proof:lower_bound}, but the following discussion outlines its main arguments.

\paragraph{Sketch of the proof in dimension 1.} 
Consider the setup where firstly $n$ points from $f$ are chosen and evaluated. Secondly, $n$ other points are sampled using rejection sampling with a proposal based only on the $n$ first points. This is related to Definition~\ref{ga:easy}.
Now, $\mathcal F_0(1,1,1/2,1)$ corresponds to one-dimensional $(1,1)$-H\"older functions which are bounded from below by $1/2$. We consider a subset of $\mathcal F_0(1,1,1/2,1)$ satisfying Assumption~\ref{ass:f}. 
Set $V_n = \{\nu=(\nu_i)_{0\leq i\leq 4n-1}\; |\; \nu_i \in \{-1,1\},;\ \sum_{i=0}^{4n-1} \nu_i=0 \}.$

Let us define the bump function $b:[0,1/(4n)]\rightarrow \mathbb{R}^+$ such that for any $\nu \in V_n$:
$$
b(x) = \begin{cases}
x, & \text{for } x \leq 1/(8n). \\
    1/(4n)-x, & \text{otherwise.}
\end{cases}
$$

\noindent
We will consider the following functions $f_{\nu}:[0,1]\rightarrow \mathbb{R}^*_+$ such that for any $\nu \in V_n$:
$$
f_{\nu}(x) = 1 + \nu_i b(x-i/(4n)),\text{ if } i/(4n) \leq x \leq (i+1)/(4n),
$$

\noindent
We note that $f_{\nu}\in \mathcal F_0(1,1,1/2,1)$, for $n$ large enough ensuring that $f_{\nu} \geq 1/2$.

An upward bump at position $i$ corresponds to $\nu_i = 1$ and a downward bump to $\nu_i = -1$. The construction presented here is analog to the one in the proof of Lemma~\ref{le:smooth_simple_fun}. The function $f_\nu$ is entirely determined by the knowledge of $\nu$. It is only possible to determine a $\nu_i$ by evaluating $f$ at some $x \in (i/(4n),(i+1)/(4n))$. So with a budget of $n$, we observe at most $n$ of the $4n$ signs in $\nu$. Among the unobserved $\nu_i$, at least $n$ are positive and $n$ are negative, because $\sum_{i=0}^{4n-1} \nu_i=0$. 
Now, we compute the loss. In the case when $Mg$ is not an envelope, the loss simply is $n$. Now let us consider the case where $Mg$ is an envelope. The loss is $n(1-1/I_{Mg})$. $Mg$ has to account for at least $n$ upward bumps at unknown positions; and the available information is insufficient to distinguish between upward and downward bumps. This results in an envelope that is not tight for the negative $\nu_i$ with unknown positions. So a necessary loss is incurred at the downward bumps corresponding to those negative $\nu_i$. This translates as $I_{Mg} - 1 \geq n \times c_sn^{-(1+s)}$, where $c_s$ is a constant only dependent on $s$, with $s=1$ in our case. 
Finally, we obtain a risk $n(1-1/I_{Mg})$ which is of order $n^{1-s}$, as seen in Lemma~\ref{le:lower_bound}.

In a nutshell, we first made a setup with more available information than in the problem of adaptive rejection sampling, from Definition~\ref{game:1}. Then we restricted the setting to some subspace of $\mathcal F_0(1,1,1/2,1)$. This led to the obtention of a lower bound on the risk for an easier setting. This implies we have displayed a lower bound for the problem of adaptive rejection sampling over $\mathcal F_0(1,1,1/2,1)$ too.
\endproof

\noindent
This theorem gives a lower bound on the minimax risk of all possible adaptive rejection sampling algorithms. Up to a $\log(n)$ factor, NNARS is minimax-optimal and the rate in the lower bound is then the minimax rate of the problem. It is remarkable that this problem admits such a fast minimax rate; the same rate as a standard rejection sampling scheme with an envelope built using the knowledge of $n$ evaluations of the target density (see Setting \ref{ga:easy} in the Appendix).

\section{Discussion}
\label{sec:discussion}
Theorem \ref{th:lower_bound} asserts that NNARS provides a minimax near-optimal solution in expectation (up to a multiplicative factor of the order of $\log(n)^{s/d}$). This result holds for all adaptive rejection sampling algorithms and densities in $\mathcal{F}_0(s,H,c_f,d)$. To the best of our knowledge, this is the first time a lower bound is proved on adaptive rejection samplers; or that an adaptive rejection sampling algorithm that achieves near-optimal performance is presented.
In order to ensure the theoretical rates mentioned in this work, the algorithm requires to know $c_f$, a positive lower bound for $f$, and the regularity constants of $f$: $s$, and $H$. Note that to achieve a near-optimal rejection rate, the precise knowledge of $s$ is required. Indeed, replacing the exponent $s$ by a smaller number will result in adding a confidence term $\hat r_{\cup_{i\leq k} \chi_i}$ to the estimator which is too large. Finally, it will result in a higher rejection rate than if one had set $s$ to the exact H\"older exponent of $f$. The assumption on $c_f$ implies in particular that $f$ does not vanish. However, as long as it remains positive, $c_f$ can be chosen arbitrarily small, and $n$ has to be taken large enough to ensure that $c_f$ is approximately larger than  $\frac{1}{\log \log n}$. When $c_f$ is not available, asymptotically taking $c_f$ of this order will offer a valid algorithm, which outputs independent samples drawn according to $f$. Moreover taking $c_f$ of this order will still result in a minimax near-optimal rejection rate. Indeed it will approximately boil down to multiplying the rejection rate by a $\log \log n$. Similarly $H$ can be taken of order $\log n$ without hindering the minimax near-optimality. Extending NNARS to non lower-bounded densities is still an open question. 

The algorithm NNARS is a perfect sampler. Since our objective is to maximize the number of i.i.d.~samples generated according to $f$, we cannot compare the algorithm with MCMC methods, which provide non-i.i.d.~samples. In our setting, they have a loss of $n$. The same argument is valid for other adaptive rejection samplers that produce correlated samples, like e.g.,~\cite{gilks1995adaptive, martino2012improved,meyer2008adaptive}.\\
Considering other perfect adaptive rejection samplers, like the ones in e.g.,~\cite{gilk1992derivative, martino2011generalization, hormann1995rejection,gorur2011concave}, their assumptions differ in nature from ours. Instead of shape constraint assumptions---like log-concavity--- which are often assumed in the quoted literature, we only assume H\"older regularity. Note that log-concavity implies H\"older regularity of order two almost everywhere. Moreover no theoretical results on the proportion of rejected samples are available for most samplers---except possibly asymptotic convergence to $0$, which is induced by our result.

PRS \citep{erraqabi2016pliable} is the only algorithm with a theoretical guarantee on the rate with the proportion of rejected samples decreasing to $0$. But it is not optimal, as explained in Section~\ref{intro}. So the near-optimal rejection rate is a major asset of the NNARS algorithm compared to the PRS algorithm. 
Besides, PRS only provides an envelope with high probability, whereas NNARS provides it with probability $1$ at any time. 
The improved performance of NNARS compared to PRS may be attributed to the use of an estimator more adapted to noiseless evaluations of $f$, and to the multiple updates of the proposal. 

\section{Experiments}
\label{sec:expes}

Let us compare NNARS numerically with Simple Rejection Sampling (SRS), PRS (\citealp{erraqabi2016pliable}), OS* (\citealp{dymetman2012algorithm}) and A* sampling (\citealp{maddison2014sampling}). The value of interest is the sampling rate corresponding to the proportion of samples produced with respect to the number of evaluations of $f$. This is equivalent to the acceptance rate in rejection sampling. Every result is an average over 10 runs with its standard deviation.




\subsection{Presentation of the experiments}
\begin{figure}
\centering
\begin{subfigure}{0.45\textwidth}
\centering
\caption{Sampling rate vs. Peakiness [Exp1]}
\label{fig:peaky}
\includegraphics[width=\linewidth]{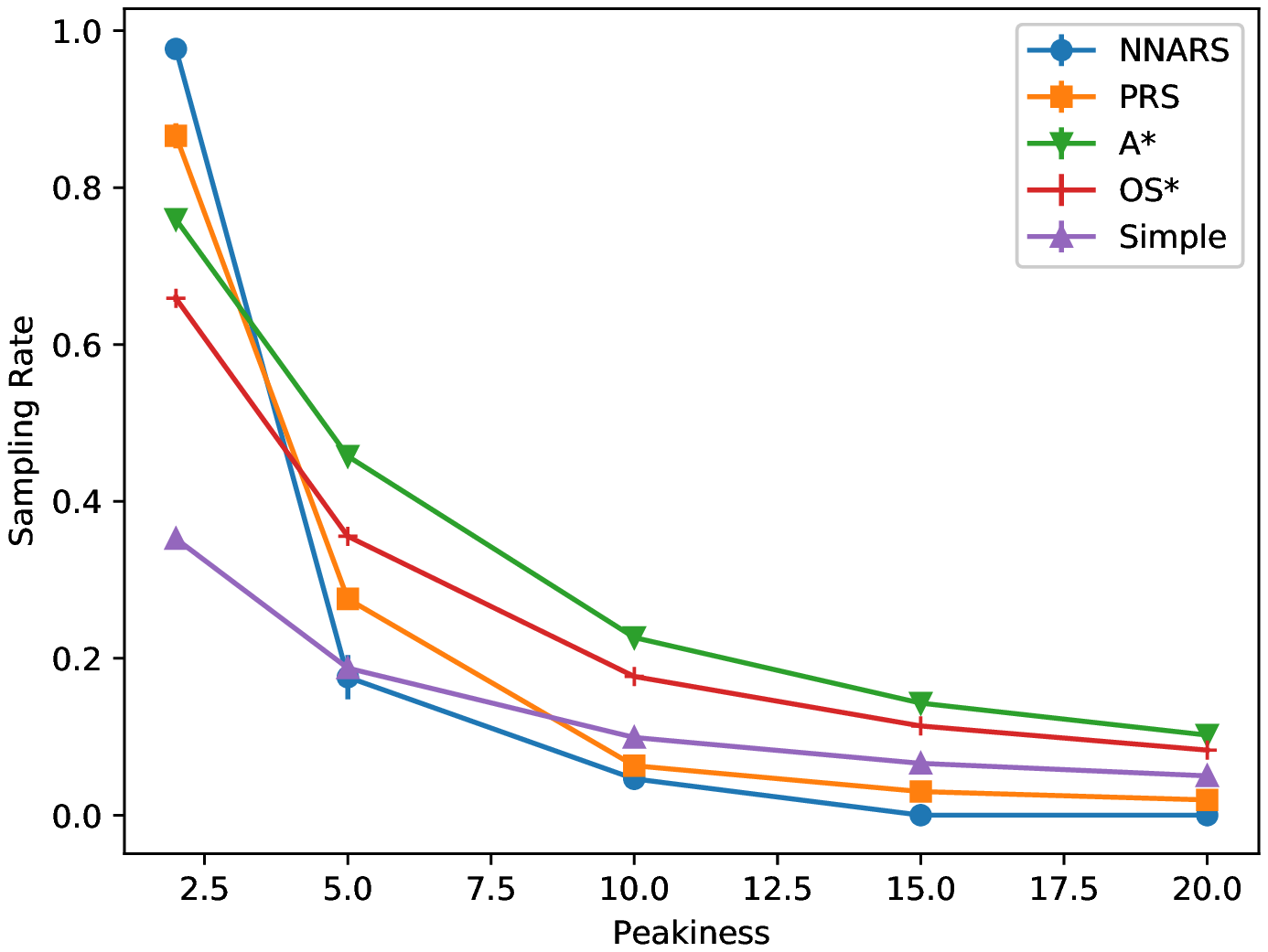}
\end{subfigure}
\begin{subfigure}{0.45\textwidth}
\centering
\caption{Sampling rate vs. Dimension [Exp2]}
\label{fig:dimension}
\includegraphics[width=\linewidth]{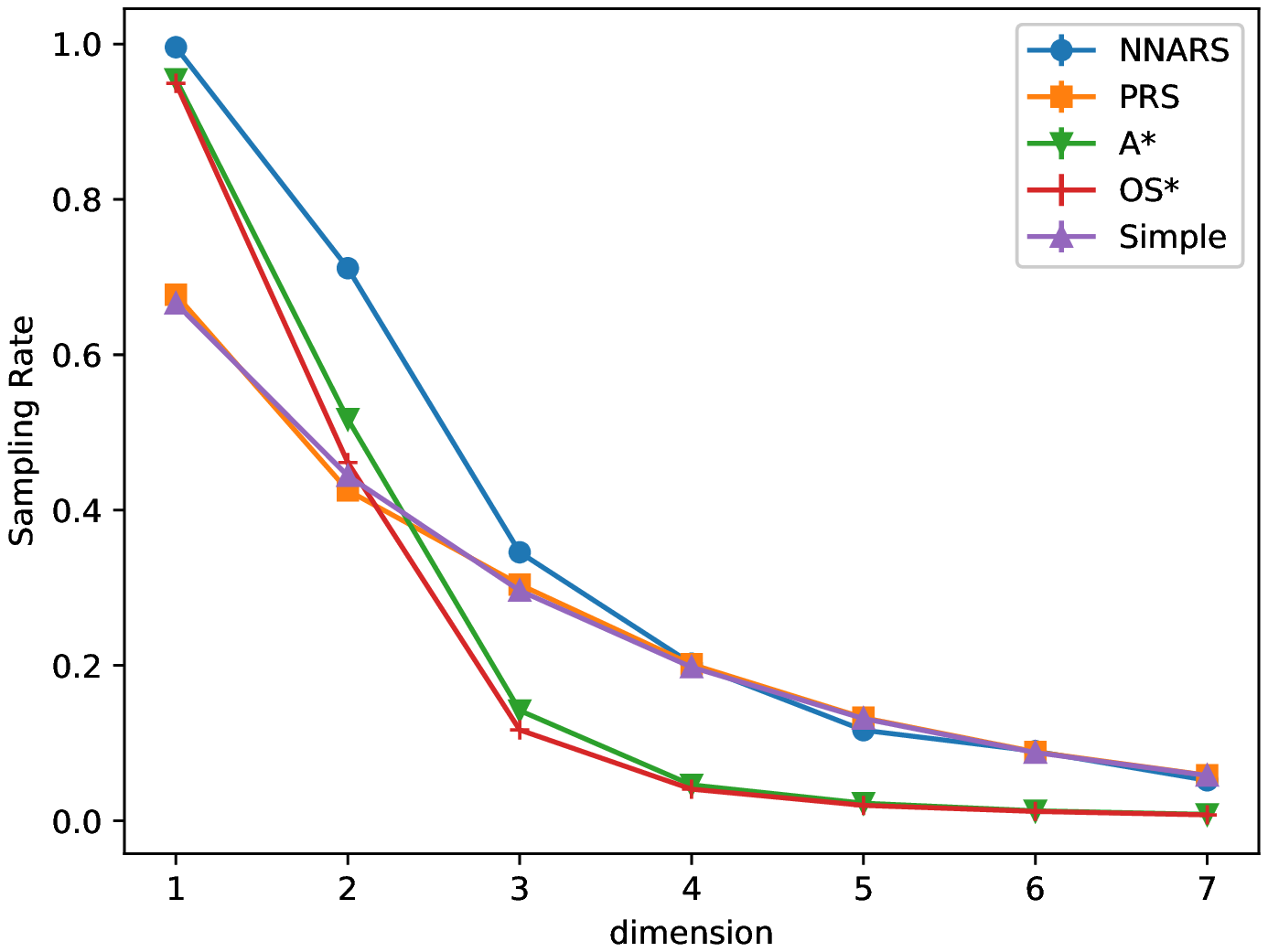}
\end{subfigure}
\caption{Empirical sampling rates for [Exp1] and [Exp2]}
\label{fig:experiment12}
\end{figure}

\begin{figure}
    \centering
\begin{subfigure}{0.45\textwidth} 
\centering
\caption{Sampling rate vs. Budget [Exp3]}
\label{fig:budget}
\includegraphics[width=\linewidth]{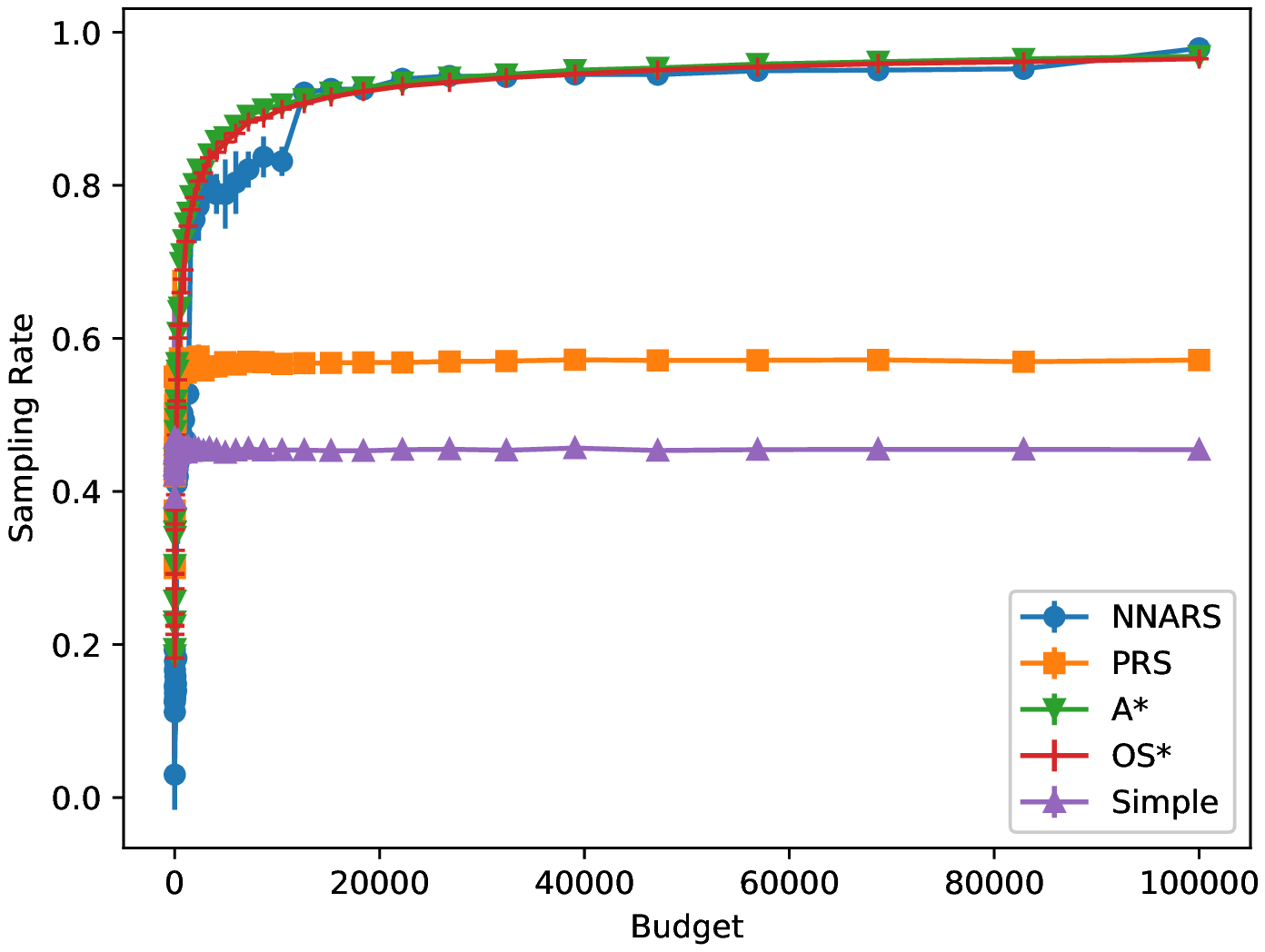}
\end{subfigure}
\begin{subfigure}{0.45\textwidth} 
\centering
\caption{Standard deviation of the sampling rate vs. budget [Exp3]}
\label{fig:budgetSTD}
\includegraphics[width=\linewidth]{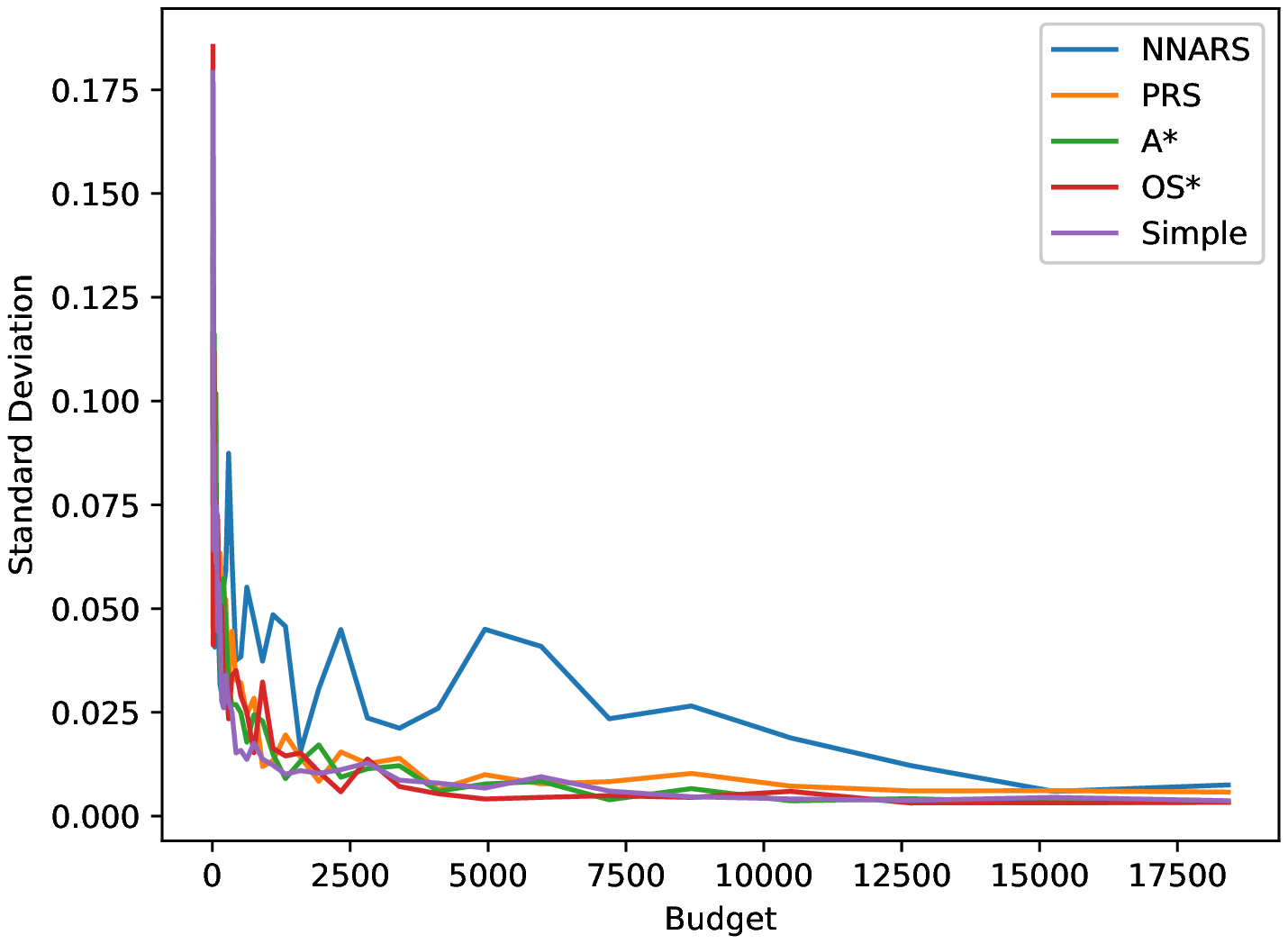}
\end{subfigure}
    \caption{Empirical sampling rates and their standard deviations for [Exp3]}
    \label{fig:experiment3}
\end{figure}

\paragraph{EXP1.}
We first consider the following target density from \cite{maddison2014sampling}: $
f(x) \propto e^{-x}/(1+x)^a,$
where $a$ is the peakiness parameter. Increasing $a$ also increases the sampling difficulty.
In Figure~\ref{fig:peaky}, PRS and NNARS both give good results for low peakiness values, but their sampling rates fall drastically as the peakiness increases. So their results are similar to SRS after a peakiness of $5.0$. On the other hand, the rates of A* and OS* sampling decrease more smoothly.
\paragraph{EXP2.}
For the next experiment, we are interested in how the method scale when the dimension increases and consider a distribution that is related to the one in~\cite{erraqabi2016pliable}: $
f(x_1,\ldots,x_d) \propto \prod_{i \in [|0,1|]^d} \left(2 + \sin\left(4\pi x_i - \frac{\pi}{2}\right)\right)
$, where $(x_1,\ldots,x_d) \in [0,1]^d$.
In Figure~\ref{fig:dimension}, we present the results for $d$ between 1 and 7. NNARS scales the best in dimension. A* and OS* have the same behaviour, while PRS and SRS share very similar results. A* and OS* start with good sampling rates, which however decrease radically when the dimension increases.
\paragraph{EXP3.}
Then, we focus on how the efficiency scales with respect to the budget. The distribution tested is: $
f(x) \propto \exp(\sin(x)),
$ with $x$ in $[0,1]$.
In Figure~\ref{fig:budget}, NNARS, A* and OS* give the best performance, reaching the asymptotic regime after 20,000 function evaluations. So NNARS is applicable in a reasonable number of evaluations. Coupled with the study of the evolution of the standard deviations in Figure~\ref{fig:budgetSTD}, we conclude that the results in the transition regime may vary, but the time to the asymptotic region is not initialization-sensitive.
\paragraph{EXP4.}
Finally, we show the efficiency of NNARS on non-synthetic data from the set in \cite{cortez2007data}. It consists of 517 observations of meteorological data used in order to predict forest fires in the north-eastern part of Portugal. The goal is to enlarge the data set. So we would like to sample artificial data points from a distribution which is close to the one which generated the data set. This target distribution is obtained in a non-parametric way, using the Epanechnikov kernel which creates a non-smooth $f$. We then apply samplers which do not use the decomposition of $f$ described in \cite{maddison2014sampling}. That is why A* and OS* sampling will not be applied. From the 13 dimensions of the dataset we work with those corresponding to Duff Moisture Code (DMC) and Drought Code (DC) and we get the sampling rates in Table~\ref{tab:fires}. NNARS clearly offers the best performance.

\begin{table}[h!]
\centering
\begin{tabular}{c|c}
n=$10^5$, 2D & sampling rate \\ \hline
NNARS                       & $45.7\% \pm 0.1\%$              \\
PRS                         & $16.0\%    \pm 0.1\%$              \\
SRS                      & $15.5\% \pm 0.1\% $            \\
\end{tabular}
\caption{Sampling rates for forest fires data [Exp4]}
\label{tab:fires}
\end{table}



\subsection{Synthesis on the numerical experiments}

The essential features of NNARS have been brought to light in the experiments presented in Figures~\ref{fig:experiment12}, \ref{fig:experiment3} and using the non-synthetic data from \cite{cortez2007data}. In particular, Figure~\ref{fig:budget} gives the evidence that the algorithm reaches good sampling rates in a relatively small number of evaluations of the target distribution. Furthermore, Figure~\ref{fig:dimension} illustrates the possibility of applying the algorithm in a multidimensional setting. In Figure~\ref{fig:peaky}, we observe that A* and OS* sampling benefit from the knowledge of the specific decomposition of $f$ needed in \cite{maddison2014sampling}. We highlight the fact that this assumption is not true in general. Besides, A* sampling requires relevant bounding and splitting strategies. We note that tuning NNARS only requires the choice of a few numerical hyperparameters. They might be chosen thanks to generic strategies like grid search. Finally, the application to forest fire data generation illustrates the great potential of NNARS for applications reaching beyond the scope of synthetic experiments.

\section{Conclusion}

In this work, we introduced an adaptive rejection sampling algorithm, which is a perfect sampler according to $f$. It offers a rejection rate of order $(\log(n)/n)^{s/d}$, if $s\leq 1$. This rejection rate is near-optimal, in the minimax sense over the class of $s$-H\"older smooth densities. Indeed, we provide the first lower bound for the adaptive rejection sampling problem, which provides a measure of the difficulty of the problem. Our algorithm matches this bound up to logarithmic terms.

In the experiments, we test our algorithm in the context of synthetic target densities and of a non-synthetic dataset. A first set of experiments shows that the behavior of the sampling rate of our algorithm is similar to that of state of the art methods, as the dimension and the budget increase.
Two of the methods used in this set of experiments require the target density to allow a specific decomposition. Therefore, these methods are neglected for the experiment which aims at generating forest fire data. In this experiment, NNARS clearly performs better than its competitors.

The extension of the NNARS algorithm to non lower-bounded densities is still an open question, as well as the development of an optimal adaptive rejection sampler, when the density's derivative is H\"older regular instead. We leave these interesting open questions for future work.
\newpage
\paragraph{Acknowledgements.} The work of A. Carpentier is partially supported by the Deutsche Forschungsgemeinschaft (DFG) Emmy Noether grant MuSyAD (CA 1488/1-1), by the DFG - 314838170, GRK 2297 MathCoRe, by the DFG GRK 2433 DAEDALUS, by the DFG CRC 1294 'Data Assimilation', Project A03, and by the UFA-DFH through the French-German Doktorandenkolleg CDFA 01-18.

\bibliographystyle{icml2018}
\small
\bibliography{ref}

\begin{thebibliography}{29}
\providecommand{\natexlab}[1]{#1}
\providecommand{\url}[1]{\texttt{#1}}
\expandafter\ifx\csname urlstyle\endcsname\relax
  \providecommand{\doi}[1]{doi: #1}\else
  \providecommand{\doi}{doi: \begingroup \urlstyle{rm}\Url}\fi

\bibitem[Andrieu et~al.(2003)Andrieu, De~Freitas, Doucet, and
  Jordan]{andrieu2003introduction}
Christophe Andrieu, Nando De~Freitas, Arnaud Doucet, and Michael~I Jordan.
\newblock An introduction to mcmc for machine learning.
\newblock \emph{Machine learning}, 50\penalty0 (1-2):\penalty0 5--43, 2003.

\bibitem[Can{\'e}vet et~al.(2016)Can{\'e}vet, Jose, and
  Fleuret]{canevet2016importance}
Olivier Can{\'e}vet, Cijo Jose, and Fran{\c{c}}ois Fleuret.
\newblock Importance sampling tree for large-scale empirical expectation.
\newblock In \emph{Proceedings of the International Conference on Machine
  Learning (ICML)}, number EPFL-CONF-218848, 2016.

\bibitem[Capp{\'e} et~al.(2008)Capp{\'e}, Douc, Guillin, Marin, and
  Robert]{cappe2008adaptive}
Olivier Capp{\'e}, Randal Douc, Arnaud Guillin, Jean-Michel Marin, and
  Christian~P Robert.
\newblock Adaptive importance sampling in general mixture classes.
\newblock \emph{Statistics and Computing}, 18\penalty0 (4):\penalty0 447--459,
  2008.

\bibitem[Cortez and Morais(2007)]{cortez2007data}
Paulo Cortez and An{\'\i}bal de Jesus~Raimundo Morais.
\newblock A data mining approach to predict forest fires using meteorological
  data.
\newblock 2007.

\bibitem[Delyon and Portier(2018)]{delyon2018efficiency}
Bernard Delyon and Fran{\c{c}}ois Portier.
\newblock Efficiency of adaptive importance sampling.
\newblock \emph{arXiv preprint arXiv:1806.00989}, 2018.

\bibitem[Devroye(1986)]{devroye1986sample}
Luc Devroye.
\newblock Sample-based non-uniform random variate generation.
\newblock In \emph{Proceedings of the 18th conference on Winter simulation},
  pages 260--265. ACM, 1986.

\bibitem[Dymetman et~al.(2012)Dymetman, Bouchard, and
  Carter]{dymetman2012algorithm}
Marc Dymetman, Guillaume Bouchard, and Simon Carter.
\newblock The os* algorithm: a joint approach to exact optimization and
  sampling.
\newblock \emph{arXiv preprint arXiv:1207.0742}, 2012.

\bibitem[Erraqabi et~al.(2016)Erraqabi, Valko, Carpentier, and
  Maillard]{erraqabi2016pliable}
Akram Erraqabi, Michal Valko, Alexandra Carpentier, and Odalric Maillard.
\newblock Pliable rejection sampling.
\newblock In \emph{International Conference on Machine Learning}, pages
  2121--2129, 2016.

\bibitem[Evans and Swartz(1998)]{evans1998random}
Michael Evans and Timothy Swartz.
\newblock Random variable generation using concavity properties of transformed
  densities.
\newblock \emph{Journal of Computational and Graphical Statistics}, 7\penalty0
  (4):\penalty0 514--528, 1998.

\bibitem[Fill(1997)]{fill1997interruptible}
James~Allen Fill.
\newblock An interruptible algorithm for perfect sampling via markov chains.
\newblock In \emph{Proceedings of the twenty-ninth annual ACM symposium on
  Theory of computing}, pages 688--695. ACM, 1997.

\bibitem[Gilks(1992)]{gilk1992derivative}
Walter~R Gilks.
\newblock Derivative-free adaptive rejection sampling for gibbs sampling,
  bayesian statistics 4, 1992.

\bibitem[Gilks and Wild(1992)]{gilks1992adaptive}
Walter~R Gilks and Pascal Wild.
\newblock Adaptive rejection sampling for gibbs sampling.
\newblock \emph{Applied Statistics}, pages 337--348, 1992.

\bibitem[Gilks et~al.(1995)Gilks, Best, and Tan]{gilks1995adaptive}
Walter~R Gilks, NG~Best, and KKC Tan.
\newblock Adaptive rejection metropolis sampling within gibbs sampling.
\newblock \emph{Applied Statistics}, pages 455--472, 1995.

\bibitem[G{\"o}r{\"u}r and Teh(2011)]{gorur2011concave}
Dilan G{\"o}r{\"u}r and Yee~Whye Teh.
\newblock Concave-convex adaptive rejection sampling.
\newblock \emph{Journal of Computational and Graphical Statistics}, 20\penalty0
  (3):\penalty0 670--691, 2011.

\bibitem[H{\"o}rmann(1995)]{hormann1995rejection}
Wolfgang H{\"o}rmann.
\newblock A rejection technique for sampling from t-concave distributions.
\newblock \emph{ACM Transactions on Mathematical Software (TOMS)}, 21\penalty0
  (2):\penalty0 182--193, 1995.

\bibitem[Maddison et~al.(2014)Maddison, Tarlow, and
  Minka]{maddison2014sampling}
Chris~J Maddison, Daniel Tarlow, and Tom Minka.
\newblock A* sampling.
\newblock In \emph{Advances in Neural Information Processing Systems}, pages
  3086--3094, 2014.

\bibitem[Martino(2017)]{martino2017parsimonious}
Luca Martino.
\newblock Parsimonious adaptive rejection sampling.
\newblock \emph{Electronics Letters}, 53\penalty0 (16):\penalty0 1115--1117,
  2017.

\bibitem[Martino and Louzada(2017)]{martino2017adaptive}
Luca Martino and Francisco Louzada.
\newblock Adaptive rejection sampling with fixed number of nodes.
\newblock \emph{Communications in Statistics-Simulation and Computation}, pages
  1--11, 2017.

\bibitem[Martino and M{\'\i}guez(2011)]{martino2011generalization}
Luca Martino and Joaqu{\'\i}n M{\'\i}guez.
\newblock A generalization of the adaptive rejection sampling algorithm.
\newblock \emph{Statistics and Computing}, 21\penalty0 (4):\penalty0 633--647,
  2011.

\bibitem[Martino et~al.(2012)Martino, Read, and Luengo]{martino2012improved}
Luca Martino, Jesse Read, and David Luengo.
\newblock Improved adaptive rejection metropolis sampling algorithms.
\newblock \emph{arXiv preprint arXiv:1205.5494}, 2012.

\bibitem[Metropolis and Ulam(1949)]{metropolis1949monte}
Nicholas Metropolis and Stanislaw Ulam.
\newblock The monte carlo method.
\newblock \emph{Journal of the American statistical association}, 44\penalty0
  (247):\penalty0 335--341, 1949.

\bibitem[Meyer et~al.(2008)Meyer, Cai, and Perron]{meyer2008adaptive}
Renate Meyer, Bo~Cai, and Fran{\c{c}}ois Perron.
\newblock Adaptive rejection metropolis sampling using lagrange interpolation
  polynomials of degree 2.
\newblock \emph{Computational Statistics \& Data Analysis}, 52\penalty0
  (7):\penalty0 3408--3423, 2008.

\bibitem[Naesseth et~al.(2017)Naesseth, Ruiz, Linderman, and
  Blei]{naesseth2017reparameterization}
Christian Naesseth, Francisco Ruiz, Scott Linderman, and David Blei.
\newblock Reparameterization gradients through acceptance-rejection sampling
  algorithms.
\newblock In \emph{Artificial Intelligence and Statistics}, pages 489--498,
  2017.

\bibitem[Naesseth et~al.(2016)Naesseth, Ruiz, Linderman, and
  Blei]{naesseth2016rejection}
Christian~A Naesseth, Francisco~JR Ruiz, Scott~W Linderman, and David~M Blei.
\newblock Rejection sampling variational inference.
\newblock \emph{arXiv preprint arXiv:1610.05683}, 2016.

\bibitem[Oh and Berger(1992)]{oh1992adaptive}
Man-Suk Oh and James~O Berger.
\newblock Adaptive importance sampling in monte carlo integration.
\newblock \emph{Journal of Statistical Computation and Simulation}, 41\penalty0
  (3-4):\penalty0 143--168, 1992.

\bibitem[Propp and Wilson(1998)]{propp1998coupling}
James Propp and David Wilson.
\newblock Coupling from the past: a user’s guide.
\newblock \emph{Microsurveys in Discrete Probability}, 41:\penalty0 181--192,
  1998.

\bibitem[Ryu and Boyd(2014)]{ryu2014adaptive}
Ernest~K Ryu and Stephen~P Boyd.
\newblock Adaptive importance sampling via stochastic convex programming.
\newblock \emph{arXiv preprint arXiv:1412.4845}, 2014.

\bibitem[Von~Neumann(1951)]{von195113}
John Von~Neumann.
\newblock 13. various techniques used in connection with random digits.
\newblock \emph{Appl. Math Ser}, 12:\penalty0 36--38, 1951.

\bibitem[Zhang(1996)]{zhang1996nonparametric}
Ping Zhang.
\newblock Nonparametric importance sampling.
\newblock \emph{Journal of the American Statistical Association}, 91\penalty0
  (435):\penalty0 1245--1253, 1996.

\end{thebibliography}
\normalsize
\newpage
\appendix


\section{Illustration of Rejection Sampling}
\begin{figure}[H]
\includegraphics[width=0.6\textwidth]{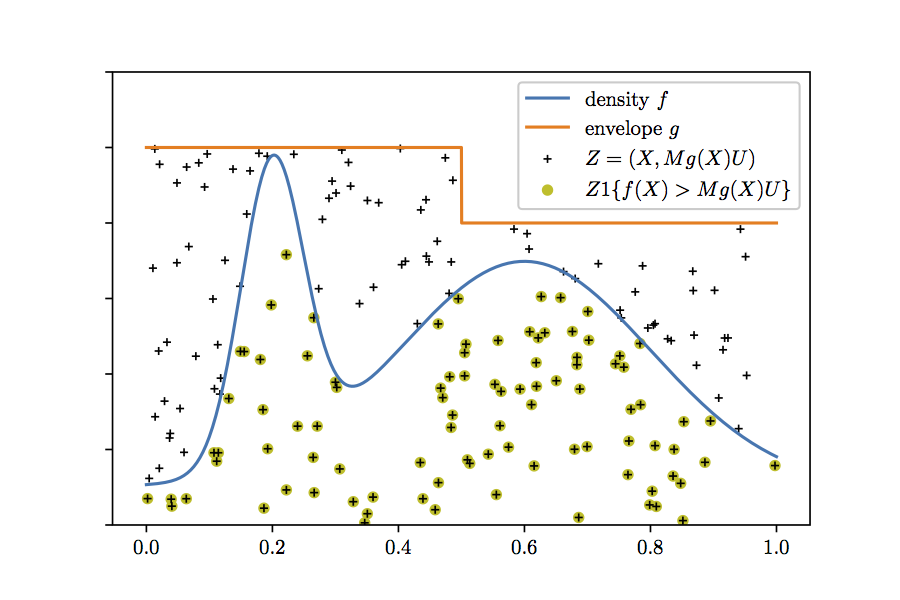}
\caption{Geometrical interpretation of Rejection Sampling}
\label{fig:geom}
\end{figure}

\noindent
\fbox{\parbox{0.985\textwidth}{ \textbf{In the following, we do not assume that $f$ is a density.} In fact ARS samplers could be given evaluations of the density multiplied by a positive constant. We prove in the sequel that as long as the resulting function satisfies Assumption \ref{ass:f}, the upper bound presented in Theorem \ref{th:speedrate} holds in this case as well as in the case when $f$ is a density. The lower bound is also proved without the assumption that $f$ is a density.}}

\section{Proof of Theorem \ref{th:ARSa}}
\label{proof:ARSa}
Let us assume that $\forall t\leq n, \forall x \in [0,1]^d, f(x)\leq M_t g_t(x)$. If $X_t$ has been drawn at time $t$, and $E_t$ denotes the event in which $X_t$ is accepted, and $\tilde \chi_j$ denotes the set of the proposal samples drawn at time $j\leq n$ and of their images by $f$, then $\forall \Omega \subset [0,1]^d$ such that $\Omega$ is Lebesgue measureable, it holds:
\begin{multline*}
  \Po_{X_t \sim {g}_t, U\sim \mathcal U_{[0,1]}}\Big(\{X_t \in \Omega\} \cap E_t \; \Big|\;\bigcup_{j<t} \chi_{j}\Big)\\
  \begin{aligned}
 &= \Po_{X_t \sim {g}_t, U\sim \mathcal U_{[0,1]}}\Big(X_t \in \Omega ;  \frac{f(X_t)}{{M}_t {g}_t(X_t)} \geq U_t \; \Big| \; \bigcup_{j<t} \chi_{j}\Big)\\
& = \int_{\Omega} \frac{f(x)}{{M}_t {g}_t(x)}  {g}_t(x) dx\\
&= \int_{\Omega} \frac{f(x)}{{M}_t}  dx,
\end{aligned}
\end{multline*}
because $U_t$ is independent from $X_t$ conditionally to $\bigcup_{j<t} \chi_{j}$.\\
Hence, since $\Po_{X_t \sim {g}_t, U\sim \mathcal U_{[0,1]}}(E_t) = {I_f}/{{M}_t} $, we have:
\begin{align*}
  \Po_{X_t \sim {g}_t, U\sim \mathcal U_{[0,1]}}\Big(X_t \in \Omega \; \Big| \; E_t ; \bigcup_{j<t}\chi_{j}\Big)&= \int_{\Omega} \frac{f(x)}{{M}_t}  \left( \frac{{M}_t}  {I_f} \right)  dx \\ &=  \int_{\Omega} \frac{f(x)}{I_f} dx.  
\end{align*}
Thus $X_t|E_t$ is distributed according to $f/I_f$ and is independent from the samples accepted before step $t$, since $X_t|E_t$ is independent from $\bigcup_{j<t}\chi_{j}$.

We have proved that the algorithm provides independent samples drawn according to the density ${f}/{I_f}$.

\section{Proof of Theorem \ref{th:speedrate}}\label{proof:speedrate}

\subsection{Approximate Nearest Neighbor Estimator}
In this subsection, we study the characteristics of the Approximate Nearest Neighbor Estimator. First, we prove a bound on the distance between the image of $x$ by the Approximate Nearest Neighbor Estimator of $f$ and $f(x)$, under the condition that $f$ satisfies Assumption \ref{ass:f}. More precisely, we prove that $\hat{f}(x)$ lies within a radius of $\hat{r}_{\chi}$ away from $f(x)$. Then we prove a high probability bound on the radius $\hat{r}_{\chi}$ under the same assumptions. This bound only depends on the probability, the number of samples, and constants of the problem. These propositions will be of use in the proof of Theorem \ref{th:speedrate}.\\

\noindent Let $\tilde N>0$, we write $C := C_{\tilde N}$ (as in Definition~\ref{def:estimator}) for simplicity.

\begin{property}
\label{pr:estimator}
Let $f$ be a positive function satisfying Assumption \ref{ass:f}.
Consider $\widetilde N$ points $\chi =\{(X_1, f(X_1)),\ldots, (X_{\widetilde N}, f(X_{\widetilde N}))\}$.\\
If $\hat{f}_{\chi}$ is the Approximate Nearest Neighbor Estimate of $f$, as defined in Definition \ref{def:estimator}, then:
$$\forall x\in[0,1]^d,\; |\hat{f}_{\chi}(x) - f(x)|\leq \hat{r}_{\chi},$$
where $\hat{r}_{\chi}$ is defined in Equation~\eqref{eq:rate} (in Definition~\ref{def:estimator}).


\end{property}

\paragraph{Proof of Proposition \ref{pr:estimator}. }~
We have that $\forall x \in [0,1]^d,$
$$\|x-X_{i(C(x))}\|_\infty \leq \|x-C(x)\|_\infty + \|C(x)-X_{i(C(x))}\|_\infty,$$
where the set $\mathcal C_{\tilde N}$ and the function $i$ are defined in Definition \ref{def:estimator}.\\
Now, $\|x-C(x)\|_\infty \leq \frac{1}{2\widetilde{N}^{\frac{1}{d}}}$ and $\|C(x)-X_{i(C(x))}\|_\infty \leq \max_{u \in \mathcal C_{\tilde N}} \|u-X_{i(u)}\|_\infty, \forall x \in [0,1]^d$ and where $\mathcal C_{\tilde N}$ is defined in Definition~\ref{def:estimator}.\\
Thus $\forall x \in [0,1]^d,$ 
\begin{align}
\|x-X_{i(C(x))}\|_\infty \leq \frac{1}{2\widetilde{N}^{\frac{1}{d}}} + \max_{u \in \mathcal C_{\tilde N}} \|u-X_{i(u)}\|_\infty,
\label{eq:sepCX}    
\end{align}
and from Assumption~\ref{ass:f} 
$$
\forall x \in [0,1]^d, \; |\hat{f}_{\chi}(x)-f(x)| \leq \hat{r}_{\chi}.
$$
\endproof

\begin{property}
\label{pr:rhat_r}
Consider the same notations and assumptions as in Proposition~\ref{pr:estimator}. Let g be a density on $[0,1]^d$ such that:
$$ \exists \;1\geq c>0 \text{ such that } \forall x \in [0,1]^d,\; c<g(x),$$
and assume that the points $X_i$ in $\chi$ are sampled in an i.i.d.~fashion according to $g$.\\
Defining $\delta_0= \frac{1}{{\widetilde N}} \exp(-\widetilde N),$ it holds for any $\delta>\delta_0$, that with probability larger than $ 1- \delta$:
$$\hat{r}_{\chi} \leq 2^s r_{\widetilde N,\delta,c}.$$
where we write $r_{\widetilde N,\delta,c} =  \displaystyle{ H
\left( \frac{\log (\widetilde N/\delta) }{c \widetilde N} \right) ^{\frac{s}{d}}}$.
\end{property}

\paragraph{Proof of Proposition \ref{pr:rhat_r}.}~
Let $\epsilon$ be a positive number smaller than $1$ such that $\epsilon^{-d}$ is an integer. We split $[0,1]^d$ in $ \frac{1}{\epsilon^d} $ hypercubes  of side-length $\epsilon$ and of centers in $\mathcal C_{\epsilon^{-d}}$. Let $I$ be one of these hypercubes, we have
$\Po(X_1 \ldots X_{\widetilde N} \notin I)\leq(1- c \epsilon^d)^{{\widetilde N}} \leq \exp{(-c  \epsilon^d {\widetilde N}}).$ \newline
So with probability larger than $1 - \exp{(-c \epsilon^d {\widetilde N}})$, at least one point has been drawn in $I$.\\
Thus $\forall x \in [0,1]^d,\; \text{ with probability larger than }1- \exp{(-c\epsilon^d {\widetilde N})}$, it holds:
$$\|x-X_{i(x)}\|_{\infty}\leq \epsilon,$$
where $i(x)= \argmin_{i \in \{1,\ldots, N\}} (\|x-X_i\|_{\infty})$ .\\
Thus $\forall x \in [0,1]^d,\; \text{ with probability larger than }  1- \delta'$,
$$\|x-X_{i(x)}\|_{\infty}\leq \left( \frac{\log (1/\delta') }{c{\widetilde N}} \right)^{\frac{1}{d}} ,$$
where $\delta' = \exp{(-c  \epsilon^d {\widetilde N})}$ (observe $\delta'>\exp(-{\widetilde N})$).\\
Thus with probability larger than $ 1- \frac{1}{\epsilon^d} \delta'$, it holds
$$\forall x \in [0,1]^d, \; \|x-X_{i(x)}\|_\infty \leq \left( \frac{\log (1/\delta') }{c{\widetilde N}} \right)^{\frac{1}{d}}.$$
With probability larger than $ 1- c{\widetilde N} \delta' > 1- \frac {c{\widetilde N}} {\log{(1/\delta')} }\delta'$, it holds
$$\forall x \in [0,1]^d, \; \|x-X_{i(x)}\|_{\infty} \leq \left( \frac{\log (1/\delta') }{c{\widetilde N}} \right)^{\frac{1}{d}} . $$\\
Hence, by letting $\delta = (c{\widetilde N}) \delta'$, with probability larger than $ 1- \delta$, 
\begin{align*}
\forall x \in [0,1]^d, \; \|x-X_{i(x)}\|_{\infty} &\leq \left( \frac{\log (c{\widetilde N}/\delta) }{c{\widetilde N}} \right)^{\frac{1}{d}} \\ &\leq  \left( \frac{\log ({\widetilde N}/\delta) }{c{\widetilde N}}\right)^{\frac{1}{d}}. 
\end{align*}
Thus $\forall \delta > c{\widetilde N} \exp(-{\widetilde N}),$
with probability larger than $ 1- \delta$, 
$$\forall x \in [0,1]^d, \; \|x-X_{i(x)}\|_\infty \leq  \left( \frac{\log ({\widetilde N}/\delta) }{c{\widetilde N}}\right)^{\frac{1}{d}},$$
and in particular, with probability larger than $1-\delta$,
$$
\max_{u \in \mathcal C_{\tilde N}} \| u - X_{i(u)} \|_\infty \leq \left( \frac{\log ({\widetilde N}/\delta) }{c{\widetilde N}}\right)^{\frac{1}{d}} .
$$

\noindent
Furthermore we have since $|\chi_{\tilde N}| = \tilde N$
$$
\frac{1}{2{\widetilde N}^{\frac{1}{d}}} \leq \max_{x \in [0,1]^d} \|x-X_i(x)\|_{\infty}.
$$

\noindent
So we also have (since $c \leq 1$  and $\log(1/\delta) \geq 1$)
$$
\frac{1}{2{\widetilde N}^{\frac{1}{d}}} \leq \left( \frac{\log ( {\widetilde N}/\delta) }{c {\widetilde N}} \right)^{\frac{1}{d}}.
$$

\noindent
Finally, from Equation~\eqref{eq:sepCX}, with probability larger than $1-\delta$, $\forall x \in [0,1]^d,$
\begin{align*}
    \|x-X_{i(C(x))}\|_\infty & \leq \frac{1}{2{\widetilde N}^{\frac{1}{d}}} + \max_{u \in \mathcal C_{\tilde N}} \| u - X_{i(u)} \|_\infty\\
    & \leq 2  \left( \frac{\log ({\widetilde N}/\delta) }{c{\widetilde N}} \right)^{\frac{1}{d}},
\end{align*}
and with probability larger than $1-\delta$,
\begin{align*}
\hat{r}_{\chi} &= H \left(\frac{1}{2{\widetilde N}^{\frac{1}{d}}} + \max_{u \in \mathcal C_{\tilde N}} \| u - X_{i(u)} \|_\infty\right)^s \\
&\leq H \left( 2 \left(\frac{\log ({\widetilde N}/\delta) }{c {\widetilde N}}\right)^{\frac{1}{d}} \right)^s = 2^s  r_{{\widetilde N},\delta,c}.
\end{align*}
\endproof

\subsection{Proof of Theorem \ref{th:speedrate}}
\label{sec:proofThSpeed}
In this subsection, we prove Theorem \ref{th:speedrate}  by first proving a high probability bound on $n-\hat{n}$. We prove this high probability bound thanks to Proposition \ref{pr:independent_sampling}, and Lemma \ref{lem:bounded_g}. Proposition \ref{pr:independent_sampling} claims that the algorithm provides independent samples drawn according to ${f}/{I_f}$, under Assumption \ref{ass:f}. The proof of Proposition \ref{pr:independent_sampling} uses Lemma \ref{lem:event} which states that $\forall x \in [0,1]^d, \; f(x) \leq {M}_k {g}_k(x)$, under the relevant assumptions. We define two events: one on which every proposal envelope until time $k+1$ is bounded from below by $\frac{3}{4} c_f$: $\mathcal{W}_{k+1}$, and the other one on which every confidence radius $\hat r_{\cup_{i \leq k} \chi_k}$ until time $k$ is upper bounded by a quantity $r_{{\widetilde N},\delta,3c_f/4}$ : $\mathcal{A}_{k,\delta}$, where $\delta$ is a confidence term (designed to be used in the high probability bound on $n-\hat{n}$). Lemma \ref{lem:bounded_g} states that the probability of the event   $\mathcal{W}_{k+1}$ conditional to the event  $\mathcal{A}_{k,\delta}$ is equal to 1, and that that the probability of the event  $\mathcal{A}_{k,\delta}$ is larger than $1-2k\widetilde{\delta}$ when ${\delta}= N/nK$. The proof of Theorem \ref{th:speedrate} uses the fact that the number of rejected samples at step $k$ on $\mathcal{A}_{k,\delta}$ is a sum of Bernoulli variables of parameter smaller than a known quantity that depends on $\widehat{\delta}$, and by applying the Bernstein inequality on this sum. The proof is then concluded by summing on $k$.\\

\noindent In this subsection, we write :
$$\hat f_k = \hat f_{\cup_{i \leq k} \chi_i},~~\text{and}~~\hat r_k = \hat r_{\cup_{i \leq k} \chi_i},$$ to ensure the simplicity of notations.
We also write:
$$r_{{\widetilde N},\delta} := r_{{\widetilde N},\delta,3c_f/4}.$$
Let us also define the events:
$$
\begin{cases}
\mathcal{W}_{k} =\{ \forall j\leq k ,\;  \forall x \in [0,1]^d, \; {g}_j(x) > \frac{3}{4} c_f\},\\
\mathcal{A}_{k,\delta} = \{ \forall j\leq k, \; \hat r_j  \leq 2^s  r_{N_j,\delta} \}.
\end{cases}$$



\begin{property} \label{pr:independent_sampling}
If Assumption~\ref{ass:f} holds, the algorithm provides independent samples drawn according to the density ${f}/{I_f}$.
\end{property}

\begin{lemma}\label{lem:event} 
Consider any $k \leq K$.
Under the assumptions made in Proposition \ref{pr:independent_sampling},
$$\forall x \in [0,1]^d, \; f(x) \leq {M}_k {g}_k(x). $$
\end{lemma}

\paragraph{Proof of Lemma \ref{lem:event}.}

$g_1$ is the uniform density and $M_1$ is taken as an upper bound on $f$. So $\forall x \in [0,1]^d$:
$$
{M_1} g_1(x) \geq f(x).
$$
 Let $k \in \{ 2, \ldots, K\}$. From Proposition~\ref{pr:estimator}:
$$\forall x\in[0,1]^d,\; |\hat{f}_{k-1}(x) - f(x)|\leq \hat{r}_{k-1}.$$
Thus, $\forall x \in [0,1]^d$:

\begin{align*}
{g}_k(x) &= \frac{\hat{f}_{k-1}(x) + \hat{r}_{k-1}}{I_{\hat{f}_{k-1}} + \hat{r}_{k-1}} \geq \frac{f(x)}{I_{\hat{f}_k-1} + \hat{r}_{k-1}} \geq \frac{f(x)}{{M}_{k}}.
\end{align*}
Hence:
$$\forall x, \; {M_k} g_k(x) \geq f(x).$$
\endproof

\paragraph{Proof of Proposition \ref{pr:independent_sampling}.} 

We have that $\forall j\leq k, ~ \forall x \in [0,1]^d, ~ f(x) \leq M_k g_k(x)$. Theorem \ref{th:ARSa} proves that the algorithm provides independent samples drawn according to the density ${f}/{I_f}$.
\endproof

\begin{lemma}\label{lem:bounded_g}
 Let $\tilde \delta = N/(nK)$. If Assumption~\ref{ass:f}  and~\ref{ass:n} hold for $n$, then
 $$
 \begin{cases}
\Po(\mathcal{W}_{1}) =1,~~\Po(\mathcal{W}_{k+1} | \mathcal{A}_{k,\widetilde{\delta}}) = 1,\\
\Po(\mathcal{A}_{k,\widetilde{\delta}}) \geq 1-2k\widetilde{\delta}.
\end{cases}
$$
\end{lemma}

\paragraph{Proof of Lemma \ref{lem:bounded_g}.}~
Since $g_1(x) =1$, $s\leq 1, c_f\leq 1$, the event $\mathcal W_1 = \{\forall x \in [0,1]^d, \; {g}_1(x) > \frac{6 }{10} c_f\}$ has probability $1$. Also by Proposition~\ref{pr:estimator} and Proposition \ref{pr:rhat_r}, the event $\mathcal{A}_{1,\widetilde{\delta}}$ has probability larger than $1-\widetilde{\delta}$.\\
Consider now that the event $\mathcal{A}_{k,\widetilde{\delta}}$ holds for a given $k \leq K$. Then by Proposition~\ref{pr:estimator} and Proposition \ref{pr:rhat_r}, it holds that for all $j \leq k$ and for all $x\in [0,1]^d$
$$|\hat f_j(x) - f(x)| \leq 2^s r_{N_j, \tilde \delta}.$$
This implies that

\begin{align*}
 {g}_{j+1}(x)
  &\geq \frac{f(x)}{ {M_{j+1}}}  \geq \frac{f(x)}{I_f + 2^{s+1} r_{N_{j}, \widetilde{\delta}}}\\
  & \geq \frac{f(x)/ I_f}{ 1 + 2^{s+1} r_{N_j, \widetilde{\delta}} / I_f } \geq \frac{f(x)}{I_f} \left( 1- 2^{s+1} \frac{r_{N_j, \widetilde{\delta}}} { I_f }\right)\\
  & \geq c_f \left( 1- 2^{s+1} \frac{r_{N_j, \widetilde{\delta}}} { I_f }\right).
\end{align*}

Hence, 
\begin{align*}
{g}_{j+1}(x) &\geq c_f \left( 1-\frac{ 2^{s+1} r_{N,\widetilde{\delta}}}{c_f} \right) \geq \frac{6}{10} c_f,
\end{align*}
where we have used $r_{N_j, \widetilde{\delta} }  \leq r_{N, \widetilde{\delta} } \leq c_f/10$ (see Assumption~\ref{ass:n}). So $\Po(\mathcal{W}_{k+1} | \mathcal{A}_{k,\widetilde{\delta}}) = 1$ and we have proved the first part of the lemma.\\

Moreover, conditional to $\mathcal{A}_{k,\widetilde{\delta}}$ we have that $g_{k+1}(x) \geq \frac{6}{10} c_f$. Then we apply Proposition~\ref{pr:estimator} and Proposition \ref{pr:rhat_r}. With probability larger than $1 - \tilde \delta$ on $\chi_k$ only, and conditional to $\mathcal{A}_{k,\widetilde{\delta}}$, it holds that for all $x\in [0,1]^d$:
$$|\hat f_{k+1}(x) - f(x)| \leq 2^s r_{N_{k+1}, \tilde \delta},$$
where we use that $\hat r_{k+1} \leq \hat r_{\chi_{k+1}}$. This implies that $\Po(\mathcal{A}_{k+1,\widetilde{\delta}} | \mathcal{A}_{k,\widetilde{\delta}}) \geq  1 - \tilde \delta$, and so for any $k \leq K$
$$\Po(\mathcal{A}_{k,\widetilde{\delta}}) \geq  (1-\tilde \delta)^k.$$
This concludes the proof since $(1-\widetilde{\delta})^k \geq 1- 2k\tilde \delta$ for $\tilde \delta \leq 1/(2K)$.
\endproof

\paragraph{Proof of Theorem \ref{th:speedrate}.} ~
Let $\tilde \delta = N/(nK)$ and $\delta = K\tilde \delta$ and let
$\hat n_k$ denote the number of accepted samples at round $k$.

\noindent
From Lemma \ref{lem:event}, we know that $\forall k \leq K $,\\
$$\forall x, \; {M_k} {g}_k(x) \geq f(x).$$
Hence, the samples accepted at step $k+1$ are independently sampled according to ${f}/{I_f}$, and $N_{k+1} -\hat{n}_{k}$, the number of rejected samples, is a sum of Bernoulli variables of parameter $1-{I_f}/{{M}_{k}}$.\\
On $\mathcal{A}_{k,\widetilde{\delta}} \cap \mathcal{W}_k$,
\begin{align*}
\frac{I_f}{{M}_{k+1}} 
&\geq \frac{I_f}{I_f + 2^{s+1} r_{N_{k}, \widetilde{\delta}}}\\
& \geq 1-\frac{ 2^{s+1} r_{N_{k},\widetilde{\delta}}}{I_f}.
\end{align*}
Thus, $
1-\frac{I_f}{{M}_{k+1}} \leq \frac{ 2^{s+1} r_{N_{k},\widetilde{\delta}}}{I_f}.
$\\
On $\mathcal{A}_{K,\widetilde{\delta}} \cap \mathcal{W}_K$, according to the Bernstein inequality, $\forall k\leq K$ the event: 
\begin{align*} 
\mathcal{V}_{k} = \Bigg\{  N_{k+1}- \hat{n}_k - \Big( 1-\frac{I_f}{{M}_{k+1}} \Big) N_{k+1} \leq \sqrt{2N_{k+1}\Big( 1-\frac{I_f}{{M}_{k+1}} \Big) \log \Big(\frac{1}{\widetilde{\delta}}\Big)} + \log \Big(\frac{1}{\widetilde{\delta}}\Big) \Bigg\}
\end{align*}
has probability larger than ${1-\widetilde{\delta}}$. \\ 
Hence on $\mathcal{A}_{K,\widetilde{\delta}} \cap \mathcal{W}_K$, $\bigcap_{k\in \{1, \ldots, K\}} \mathcal{V}_k$ has probability $1-K \widetilde{\delta}$.\\
Consequently, since $\mathcal{A}_{K,\widetilde{\delta}} \cap \mathcal{W}_K$ has probability larger than $1-K \widetilde{\delta}$ according to Lemma \ref{lem:event},  $\bigcap_{k\in \{1, \ldots, K\}} \mathcal{V}_k \cap \mathcal{A}_{K,\widetilde{\delta}} \cap \mathcal{W}_K$ has probability larger than ${1- 2 K \widetilde{\delta}}$.\\
On $\mathcal{V}_k \cap \mathcal{A}_{K,\widetilde{\delta}} \cap \mathcal{W}_K$, 
\begin{align*} 
N_{k+1}- \hat{n}_k - \frac{ 2^{s+1} r_{N_{k},\widetilde{\delta}}}{I_f}N_{k+1} \leq \sqrt{2N_{k+1} \frac{ 2^s r_{N_k,\widetilde{\delta}}}{I_f} \log \left(\frac{1}{\widetilde{\delta}}\right)} + \log \left(\frac{1}{\widetilde{\delta}}\right) 
\end{align*}
(and we know from Proposition \ref{pr:independent_sampling}, that on $\mathcal{V}_k \cap \mathcal{A}_{K,\widetilde{\delta}} \cap \mathcal{W}_K$, we also have that the drawn samples are independently drawn according to ${f}/{I_f}$).\\
Hence on $\bigcap_k{\mathcal{V}_k}\cap  \mathcal{A}_{K,\widetilde{\delta}} \cap \mathcal{W}_K$, which has probability larger than $1-2K \widetilde{\delta} :=1 - 2\delta $:
\begin{align*} \sum \limits_1^{K-1} \left( N_{k+1} - \hat{n}_k - \frac{ 2^{s+1} r_{N_k,\widetilde{\delta}}}{I_f}N_{k+1} \right)\leq \sum \limits_1^{K-1} \left( \sqrt{2N_{k+1} \frac{ 2^{s+1} r_{N_k,\widetilde{\delta}}}{I_f} \log(\frac{1}{\widetilde{\delta}}) }  \right) + K \log \left(\frac{1}{\widetilde{\delta}}\right),
\end{align*}
i.e:
\begin{align*}
 n - \hat{n}  \leq \underbrace{ \frac{2^{s+1}} {I_f}\sum \limits_1^{K-1} \left( r_{N_k,\widetilde{\delta}}N_{k+1} \right) + 4 \sqrt{\frac{\log(\frac{1}{\widetilde{\delta}})}{I_f}} \sum \limits_1^{K-1} \left( \sqrt{N_{k+1} r_{N_k,\widetilde{\delta}}}  \right)}_\text{(1)}+ \underbrace{K \log \left(\frac{1}{\widetilde{\delta}}\right)}_\text{(2)} 
+\underbrace{N}_\text{(3)}.
\end{align*}


Hence, if $\beta = \frac{2^{s+1}} {I_f} +  4\sqrt{\frac{\log(\frac{1}{\widetilde{\delta}})}{ I_f}} $, and $\tilde C = H (10/(6c_f))^{s/d}$,
\begin{align*}
(1) &\leq \beta\Big[ \sum \limits_1^{K-1} \left( r_{N_k,\widetilde{\delta}}N_{k+1} \right) + K\Big]\\
 & \leq \beta \sum \limits_1^{K-1} \left( \widetilde{C} \log\Big(\frac{p^k N}{\tilde \delta}\Big)^{ s/d} (p^kN)^{1-s/d} \right) + K\beta \\
 & \leq \beta   \widetilde{C} \left(\log \left( \frac{n}{ \widetilde{\delta}}\right) \right)^{s/d}N^{1-s/d}\left( \frac{p^{(1-s/d)K}-1}{p-1}\right) + K\beta\\
 &\leq  \frac{\beta   \widetilde{C}} {p-1} \left(\log \left(\frac{n}{ \widetilde{\delta}} \right)\right)^{s/d} n^{1-s/d} + K\beta,
\end{align*}
and
\begin{align*}
 (2)&=K \log \Big(\frac{K}{\tilde \delta}\Big)\\
 & = \log_p \left( \frac {n}{N} \right) \log \left (\frac{\log_p(n/N)}{\widetilde{\delta}} \right) .
\end{align*}
We have proved that if the assumptions of Theorem \ref{th:speedrate} are satisfied, with probability $1- 2 \delta$, 
\begin{align*}
n - \hat{n} &\leq 
\frac{\left(\frac{2^{s+1}} {I_f} + 4\sqrt{\frac{\log(\frac{1}{\widetilde{\delta}})}{ I_f}}\right)  \widetilde{C}} {p-1} \left(\log \left(\frac{n}{2 \widetilde{\delta}} \right)\right)^{s/d} n^{1-s/d} \\ + 
& ~~~ \log_p \left( \frac {n}{N} \right) \log \left (\frac{\log_p(n/N)}{\widetilde{\delta}} \right) +N + \left(\frac{2^{s+1}} {I_f} + 4\sqrt{\frac{\log(\frac{1}{\widetilde{\delta}})}{ I_f}}\right)  \log_p \left( \frac {n}{N} \right).
\end{align*}

Finally, the proof is finished following a few strings of inequalities and taking the expected value. The following reminders may help:\\
$d\geq 1$; $s\leq 1$; $c_f \leq 1$.\\
$p = \ceil{3/(2c_f)}$; $\tilde C = H(10/(6c_f))^{s/d}$; $\delta = N/n = K \tilde \delta$; $K = \ceil{\log_p(n/N)}$ and $N = \ceil{2(10H)^{d/s} \log(n)c_f^{-1-d/s}}$.

In particular, we have:
$I_f \geq c_f$; $p-1 \geq 1/(2c_f)$; $\log p \geq \log 2$ and $1/\tilde \delta \leq 5n^2$.

\endproof

\section{Proof of Theorem \ref{th:lower_bound}.}
\label{proof:lower_bound}
\subsection{Setting}
Let us introduce two different settings:
\begin{setting}\label{ga:easy} {\bf (Class of Rejection Samplers with Access to Multiple Evaluations of the density (RSAME))}$\;$\\
A sampler belongs to the RSAME class if it follows the following steps:
\begin{itemize}
\item For each step  $t \in \{1\ldots n\}$:\\
Choose a distribution $\mathcal{D}_t$ on $\mathbb{R}$, depending on $\big((Y_1, f(Y_1)) \ldots (Y_{t-1},f(Y_{t-1}))\big)$. Draw $Y_t$ according to $\mathcal{D}_t$.
\item Choose a density $g$ and a positive constant $M$ depending on \\$\big((Y_1, f(Y_1)) \ldots (Y_{n},f(Y_{n}))\big)$, and sample $Z$ by performing one Rejection Sampling Step($f,M,g$).
\end{itemize}
{\bf Objective :} The objective of a RSAME sampler is to sample one point according to a normalized version of $f$.\\
{\bf Loss :}  The loss of a RSAME sampler is defined as follows : $$L_n'=  n( 1 - \mathbf 1\{Z \text{ is accepted } \}\mathbf 1\{f \leq M g\}).$$
{\bf Strategy :} A strategy $\mathfrak{s}'$ consists of the choice of $\mathcal{D}_t$ depending on \\
$\big((Y_1, f(Y_1)) \ldots (Y_{t-1},f(Y_{t-1}))\big)$, and of the choice of $M,g$ depending on \\$\big((Y_1, f(Y_1)) \ldots (Y_n,f(Y_n))\big).$ Denote $\mathfrak{S}'$ the set of  strategies for this setting.
\end{setting}

\begin{setting}\label{ga:hard}  {\bf (Class of Adaptive Rejection Samplers (ARS))}$\;$\\
A sampler belongs to the ARS class if, at each step t $\in \{1\ldots n\}$: it
\begin{itemize}
\item Chooses a density $g_t$, and  a positive constant $\;M_t$, depending only on \\$\Big\{(X_1, f(X_1)), \ldots, (X_{t-1},f(X_{t-1}))\Big\}$.
\item Samples $X_t$ by performing rejection sampling on the target function $f$ using $M_t$ and $g_t$ as the rejection constant and the proposal. Store $X_t$ in $\mathcal{S}$ if it is accepted.
\end{itemize}
{\bf Objective :} The objective of an ARS sampler is to sample i.i.d.~points according to a normalized version of $f$\\
{\bf Loss :} The loss of an ARS sampler is defined as follows :  $L_n=  n-\#\mathcal S\mathbf 1\{\forall t\leq n, f \leq M_tg_t\}$.\\
{\bf Strategy :} A strategy $\mathfrak{s}$ consists of the choice of $M_t,g_t$ depending on\\ $\big((X_1, f(X_1)) \ldots (X_{t-1},f(X_{t-1}))\big)$. Denote $\mathfrak{S}$ the set of strategies for this setting.
\end{setting}

For the class of samplers defined in Setting \ref{ga:hard} (and similarly for Setting \ref{ga:easy}) we call value of the class the quantity $\inf_{\mathfrak{s} \in \mathfrak{S}} \sup_{f \in \mathcal{F}_0} \E^{(\mathfrak{s},f)}(L_n)$, where the symbol $\E^{(\mathfrak{s},f)}$ denotes the expectation
with respect to all relevant random variables, when those are generated by a sampler of the relevant class, using function $f$ and strategy $\mathfrak{s}$; and $\mathcal{F}_0$ denotes the set of functions satisfying Assumption \ref{ass:f}.
\subsection{Setting Comparison}
\begin{lemma}
\label{le:game_comparison}
The value of the class defined in Setting~\ref{ga:easy} is smaller than the value of the class defined in Setting~\ref{ga:hard}:
\[
\inf_{\mathfrak{s}' \in \mathfrak{S}'} \sup_{f \in \mathcal{F}_0} \E^{(\mathfrak{s}',f)}(L'_n)
\leq \inf_{\mathfrak{s} \in \mathfrak{S}} \sup_{f \in \mathcal{F}_0} \E^{(\mathfrak{s},f)}(L_n).
\]
In other terms, Setting \ref{ga:easy} is easier than Setting \ref{ga:hard}
\end{lemma}

\paragraph{Proof of Lemma \ref{le:game_comparison}.}
For any given strategy $\mathfrak{s}$ designed for Setting~\ref{ga:hard} that chooses $(g_i,M_i)$ to generate $X_i$, consider the associated strategies $\mathfrak{s}'_1,\ldots,\mathfrak{s}'_n$ for
Setting~\ref{ga:easy} consisting of:
\begin{enumerate}
\item Generating $Y_1,\ldots,Y_{n-1}$ from the same probability distributions as $X_1,\ldots,X_{n-1}$
generated for Setting~\ref{ga:hard} using strategy $\mathfrak{s}$; this is a valid choice since the distribution $\mathcal{D}_t$ of $X_t$ only depends on $\big((X_1,f(X_1),\ldots,(X_{t-1}, f(X_{t-1})) \big)$.
\item Using $(g_i,M_i)$, given by step $i$ of strategy $\mathfrak{s}$
applied to $\big((Y_1,f(Y_1),\ldots,(Y_{i-1}, f(Y_{i-1})) \big)$, in order to sample $Z$ by rejection sampling. It is still a valid choice, which actually discards the information of
$\big((Y_i,f(Y_i),\ldots,(Y_{n-1}, f(Y_{n-1})) \big)$. 
\end{enumerate}
 Then, we have for any $f \in \mathcal{F}_0$:
\begin{align*}
\E^{(\mathfrak{s},f)}(L_n)
 & = n- \E^{(\mathfrak{s},f)}(\#\mathcal S\mathbf 1\{\forall t\leq n, f \leq M_tg_t\})\\
 & = n - \E^{(\mathfrak{s},f)}\big(\sum_{i=1}^n \mathbf 1\{X_i \text{ is accepted }\}\times \mathbf 1\{\forall t\leq n, f \leq M_tg_t\}\big)\\
 & \geq n - \E^{(\mathfrak{s},f)}\big(\sum_{i=1}^n \mathbf 1\{X_i \text{ is accepted }\}\times \mathbf 1\{f \leq M_ig_i\}\big)\\
 & = \sum_{i=1}^n \E^{(\mathfrak{s}_i,f)}\Big(  1 - \mathbf 1\{X_i \text{ is accepted }\} \times \mathbf 1\{f \leq M_ig_i\} \Big)\\
 & = \frac{1}{n} \sum_{i=1}^n \E^{(\mathfrak{s}_i,f)}(L'_n).
\end{align*}
Hence, there exists at least one strategy amongst $\mathfrak{s}'_1,\ldots,\mathfrak{s}'_n$ that reaches an expected loss in Setting~\ref{ga:easy} lower than that of strategy $\mathfrak{s}$ in Setting~\ref{ga:hard}.


\endproof

\subsection{Lower Bound for Setting \ref{ga:easy}}
\begin{lemma}\label{le:lower_bound}
\begin{align*}
\inf_{\mathfrak{s}'\in \mathfrak{S}'} \sup_{f \in \mathcal{F}_0}
\E_f(L_n'(\mathfrak{s}'))   \geq 3^{-1}2^{-1-3s-2d}5^{-s/d}n^{1-s/d},
\end{align*}
for n large enough.
\end{lemma}

\noindent
The Theorem is a direct consequence of Lemmas~\ref{le:game_comparison} and \ref{le:lower_bound}. We use Lemma~\ref{le:smooth_simple_fun} to prove Lemma~\ref{le:lower_bound}.

\begin{lemma}
\label{le:smooth_simple_fun}
Let \begin{align}
 \label{eq:F'1}
 \begin{split}
     \mathcal{F}_1'=&\Bigg\{ f \text{ s.t. } \forall u=(k_1, \ldots k_d) \in \{0,1,\ldots,a_{n,d}-1\}^d,\\ 
& \text{either } \forall x \in H_{u} := \Big[\frac{k_1}{a_{n,d}},\frac{k_1+1}{a_{n,d}} \Big]\times \ldots \times \Big[\frac{k_d}{a_{n,d}},\frac{k_d+1}{a_{n,d}}\Big],\\
                 &f(x) =
     \phi^{+}\Big(x-\frac{u}{a_{n,d}}\Big) ,\\ & \text{or }  \forall x \in H_{u},
                  f(x) = \phi^{-}\Big(x-\frac{u}{a_{n,d}}\Big) 
 \Bigg\},
 \end{split}
 \end{align}\\
 where:
\begin{align*}
&a_{n,d}= \min \{2p\in \mathbb{N}; 2p\geq (4n)^{\frac{1}{d}}\},\\
&\phi^+(x)= 1 + (2a_{n,d})^{-s} -
\left\| x-\frac{1}{2a_{n,d}} \mathbf{I} \right\|_\infty^s,\\
 &(\text{with }  \mathbf{I} \text{ denoting the unit vector}),\\
&\phi^-(x) = 2 -\phi^+(x).
\end{align*}
Then any function in $\mathcal{F}'_1$ is s-H\"older-smooth.
\end{lemma}

\begin{remark}
If $d = 1$, 
\begin{align}
 \begin{split}
     \mathcal{F}_1'=&\Bigg\{ f \text{ s.t. } \forall i \in \{0,1,\ldots,4n-1\},\\ 
& \text{either } \forall x \in H_{i} := \Big[\frac{i}{4n},\frac{i+1}{4n} \Big], \ f(x) =
     \phi^{+}\Big(x-\frac{i}{4n}\Big) ,\\ & \text{or }  \forall x \in H_{i},
                  f(x) = \phi^{-}\Big(x-\frac{i}{4n}\Big) 
 \Bigg\},
 \end{split}
 \end{align}
with $\forall x \in [0,1/(4n)]$
$$
\phi^+(x)= 1+ (8n)^{-s} -\left| x-\frac{1}{8n} \right|^s,
$$
and
$$
\phi^-(x)= 1 - (8n)^{-s} +\left| x-\frac{1}{8n} \right|^s.
$$
\end{remark}

\paragraph{Proof of Lemma \ref{le:smooth_simple_fun}.} Let us first prove that $\phi^+$ is s-smooth.
 $| \phi^+(x) - \phi^+(y) | =  |\| y-\frac{1}{2a_{n,d}} \mathbf{I} \|_{\infty}^s - \| x-\frac{1}{2a_{n,d}} \mathbf{I} \|_{\infty}^s | \leq \| x- y \|_{\infty}^s$. \\
 It is straightforward to see that $\phi^-$ is also s-smooth and that all $f\in \mathcal{F}'_1 $ are also s-smooth.
\endproof

\paragraph{Proof of Lemma \ref{le:lower_bound}.} 
Let us consider Setting \ref{ga:easy} on a subset of functions of $\mathcal{F}_0$. Let
 \begin{align*}
 \mathcal{F}_1 =  \mathcal{F}_{int} \cap \mathcal{F}_1',
 \end{align*}
 where $\mathcal{F}_1'$ is defined in Equation~\eqref{eq:F'1} and
 \begin{align*}
 \mathcal{F}_{int} = \left\{f, \int_0^1 f = 1 \right\}. 
 \end{align*}
\noindent
And $\mathcal{F}_1$ is not empty since $a_{n,d}$ defined in equation~\eqref{eq:F'1} is even.
Since $\mathcal{F}_1 \subset \mathcal{F}_0$ by application of Lemma~\ref{le:smooth_simple_fun},
$$ \inf_{\mathfrak{s}'\in \mathfrak{S}'} \sup_{f \in \mathcal{F}_0}
\E_f(L_n'(\mathfrak{s}'))   \geq \inf_{\mathfrak{s}'\in \mathfrak{S}'} \sup_{f \in \mathcal{F}_1}
\E_f(L_n'(\mathfrak{s}')) .$$
We first note that: $$ \inf_{\mathfrak{s}'\in \mathfrak{S}'} \sup_{f \in \mathcal{F}_0}
\E(L_n'(\mathfrak{s}'))   \geq \inf_{\mathfrak{s}'\in \mathfrak{S}'} \E_{f  \sim \mathcal{D}_{\mathcal{F}_1}}(L_n'(\mathfrak{s}')),$$
where $\mathcal{D}_{\mathcal{F}_1}$ is the distribution such that for any $F$, $\mathbb{P}_{f\sim\mathcal{D}_{\mathcal{F}_1}}(f = F) = \frac{\mathbf{1}\{F\in \mathcal{F}_1\}}{\# \mathcal{F}_1}$. A hypercube will refer to a $H_u$ as defined in Equation~\eqref{eq:F'1}.

We also note that the choice of $M,g$ where $M$ is a multiplicative constant and $g$ is a density is equivalent to the choice of a positive function $G$, where $G=Mg$, or $M=I_G$ and $g= \frac{G}{I_G}$.

\noindent
Furthermore a strategy $\mathfrak{s}'$ for this setting is the combination of three strategies:
\begin{enumerate}
\item $\mathfrak{s}'_1$: The strategy to choose $Y_1 \ldots Y_n$,
\item $\mathfrak{s}'_{2}$: The strategy to choose $G$.
\end{enumerate}
For the first step, let us fix a strategy $\mathfrak{s}'_1$. 
Let $f_1$ be a realization of $\mathcal{D}_{\mathcal{F}_1}$. Then by application of strategy $\mathfrak{s}'_1$, $Y_1, \ldots, Y_n$ are drawn. Then the evaluations $f_1(Y_1), \ldots, f_1(Y_n)$ are obtained. Now let $u_1, \ldots u_n$ be the indices such that $H_{u_1} \ldots H_{u_n}$ are the hypercubes where $Y_1 \in H_{u_1}, \ldots, Y_n\in H_{u_n}$.

Let us define the restricted set $\mathcal{F}_{1| f_1} = \{f \in \mathcal{F}_1\; \text{ and }\; f(Y_1) = f_1(Y_1),\ldots, f(Y_n)=f_1(Y_n)\}$. And we consider the distribution $\mathcal{D}_{\mathcal{F}_{1| f_1}}$ such that $\Po_{f\sim\mathcal{D}_{\mathcal{F}_{1| f_1}}}(f = F) = \frac{\mathbf{1}\{F\in \mathcal{F}_1\}}{\# \mathcal{F}_1}$.


\noindent
In a second step, let us fix a strategy $\mathfrak{s}'_2$. This defines a distribution $\mathcal D_G$ corresponding to the choice of $G$. By the law of total expectation, we have:
\begin{align*}
  \E_{f  \sim \mathcal{D}_{\mathcal{F}_1}}(L_n'(\mathfrak{s}'))  
  &= \E_{f_1 \sim  \mathcal{D}_{\mathcal{F}_1}}\E_{G\sim \mathcal D_G}\Big[ \E_{f \sim \mathcal{D}_{\mathcal{F}_{1| f_1}}} \Big(L_n'\big(\mathfrak{s}'\big)|(Y_1,f_1(Y_1)), \ldots (Y_n,f_1(Y_n)), G\Big)\\
  &~~~~~~~~~~~~~~~~~~~~~~~~~~~~~~~|(Y_1,f_1(Y_1)), \ldots (Y_n,f_1(Y_n))\Big]\\
 &=\E_{f_1 \sim  \mathcal{D}_{\mathcal{F}_1}} \E_{G\sim \mathcal D_G}\Big[\E_{f \sim \mathcal{D}_{\mathcal{F}_{1| f_1}}} \Big(L_n'\big(\mathfrak{s}'\big)|(Y_1,f(Y_1)), \ldots (Y_n,f(Y_n)), G\Big)\\
 &~~~~~~~~~~~~~~~~~~~~~~~~~~~~~~~|(Y_1,f_1(Y_1)), \ldots (Y_n,f_1(Y_n))\Big].
\end{align*}


\noindent
We can write:
\begin{align*}
& \E_{f \sim \mathcal{D}_{\mathcal{F}_{1| f_1}}} \left( L_n'\left( \mathfrak{s}'\right)| (Y_1,f(Y_1)), \ldots (Y_n,f(Y_n)), G \right)  \\ 
& =  \E_{f \sim \mathcal{D}_{\mathcal{F}_{1| f_1}}} \Bigg[\mathbf{1} \left\{\exists u \notin \{u_1 \ldots u_n\},\; \exists x \in H_u:  G(x) < f(x)\right\} \times n\\
&~~~ + \mathbf{1} \left\{ \forall u \notin \{u_1 \ldots u_n\},\; \forall x \in H_u:  G(x) \geq f(x) \right\} \times n \left(1- \frac{1}{1+ \| f- G\|_1}\right)\\
&~~~~~~~~~~~~ \Big| ~ (Y_1,f(Y_1)) \ldots (Y_n, f(Y_n)), G\Bigg]\\
& \geq \E_{f \sim \mathcal{D}_{\mathcal{F}_{1| f_1}}} \left(1- \frac{1}{1+ \| f- G\|_1} \Big| G\right) \times n.
\end{align*}

\noindent
Now, since for any $x \geq 0$,

$$
1-\frac{1}{1+x}\geq\frac{1}{2} (1 \wedge x),
$$
we have
\begin{align*}
 \E_{f \sim \mathcal{D}_{\mathcal{F}_{1| f_1}}} &\left( L_n'\left( \mathfrak{s}'\right)| (Y_1,f(Y_1)), \ldots (Y_n,f(Y_n)), G \right)  \\
&~~~~~~~~~~~~~~~~~~~~~~~~ \geq \frac{1}{2} \E_{f \sim \mathcal{D}_{\mathcal{F}_{1| f_1}}} \left(\| f- G\|_1\wedge 1 \Big| G\right)  \times n \\
&~~~~~~~~~~~~~~~~~~~~~~~~\geq \frac{1}{2} \E_{f \sim \mathcal{D}_{\mathcal{F}_{1| f_1}}} \left(\big\| |f- G|\wedge 1\big\|_1 \Big| G\right)  \times n\\
&~~~~~~~~~~~~~~~~~~~~~~~~\geq \frac{1}{2} \left\|\E_{f \sim \mathcal{D}_{\mathcal{F}_{1| f_1}}} \left( |f- G|\wedge 1 \Big| G\right) \right\|_1 \times n\\
&~~~~~~~~~~~~~~~~~~~~~~~~\geq \frac{1}{2} \left\|\E_{f \sim \mathcal{D}_{\mathcal{F}_{1| f_1}}} \left( [ |f- G| \wedge 1] (1-\mathbf{1}\{\cup_{i=1}^n H_{u_i}\}) \Big| G\right)  \right\|_1 \times n.
\end{align*}

\noindent
For any $u\neq u_1\ldots u_n$, $\forall x \in H_u,$ since any realization from $\mathcal{D}_{\mathcal{F}_{1| f_1}}$ is in $\mathcal F_{int}$ almost surely,

\begin{align*}
 \E_{f \sim \mathcal{D}_{\mathcal{F}_{1| f_1}}}\left(|f(x) - G(x)| \wedge 1\right) \geq \frac{1}{3}\Bigg[&\left(\left|\phi^+\left(x-\frac{u}{a_{n,d}}\right)-G(x)\right|\wedge 1\right)
 \\
 & + \left(\left|\phi^-\left(x-\frac{u}{a_{n,d}}\right)-G(x)\right|\wedge 1\right)\Bigg]
\end{align*}
And, for any $u \notin \{u_1,\ldots,u_n\}$, and for any $x \in H_u$:
\begin{align*}
&\left(\left|\phi^+\left(x-\frac{u}{a_{n,d}}\right)-G(x)\right| \wedge 1\right) + \left(\left|\phi^-\left(x-\frac{u}{a_{n,d}}\right)-G(x)\right| \wedge 1\right) \\
&~~~~~~~~~~~~~~~~~~~~~~~~\geq \left(\left|\phi^+\left(x-\frac{u}{a_{n,d}}\right)-G(x)\right| \wedge 1\right) \vee \left(\left|\phi^-\left(x-\frac{u}{a_{n,d}}\right)-G(x)\right| \wedge 1\right) \\
&~~~~~~~~~~~~~~~~~~~~~~~~\geq \min_\theta \left[\left(\left|\phi^+\left(x-\frac{u}{a_{n,d}}\right)-\theta\right| \wedge 1\right) \vee \left(\left|\phi^-\left(x-\frac{u}{a_{n,d}}\right)-\theta\right| \wedge 1\right)\right] \\
&~~~~~~~~~~~~~~~~~~~~~~~~=  \left(\left|\phi^+\left(x-\frac{u}{a_{n,d}}\right)-1\right|\wedge 1\right) \vee \left(\left|\phi^-\left(x-\frac{u}{a_{n,d}}\right)-1\right|\wedge 1\right).
\end{align*}

\noindent
And since $|\phi^+ - 1| = |\phi^- - 1|$, we end up with:
\begin{align*}
\left(\left|\phi^+\left(x-\frac{u}{a_{n,d}}\right)-G(x)\right|\wedge 1\right) +& \left(\left|\phi^-\left(x-\frac{u}{a_{n,d}}\right)-G(x)\right| \wedge 1\right) \\
&\geq \left|\phi^+\left(x-\frac{u}{a_{n,d}}\right)-1\right| \wedge 1 = \phi^+\left(x-\frac{u}{a_{n,d}}\right)-1.
\end{align*}



\noindent
So
\begin{align*}
 \E_{f \sim \mathcal{D}_{\mathcal{F}_{1| f_1}}} &\left( L_n'\left( \mathfrak{s}'\right)| (Y_1,f(Y_1)), \ldots (Y_n,f(Y_n)), G \right)  \\
&~~~~~~~~~~~~~~~~ \geq \frac{1}{6} \sum_{u\neq u_1, \ldots u_n} \int_{H_u} \left(\phi^+\left(x-\frac{u}{a_{n,d}}\right)-1\right)dx.
\end{align*}




\noindent
And 
\begin{align*}
\sum_{u\neq u_1, \ldots u_n} \int_{H_u} \left(\phi^+\left(x-\frac{u}{a_{n,d}}\right)-1\right)dx &\geq (a_{n,d}^d-n) \int_{[0,1/a_{n,d}]^d} \left(\phi^+(x) - 1\right) dx \\   
&\geq (a_{n,d}^d-n) \int_{[1/(4a_{n,d}),3/(4a_{n,d})]^d} \left(\phi^+(x) - 1\right) dx.
\end{align*}





\noindent
Now, for any $x\in[1/(4a_{n,d}),3/(4a_{n,d})]^d$, we have $\phi^+(x)-1\geq (4a_{n,d})^{-s}$.

\noindent
Then
\begin{align*}
    \sum_{u\neq u_1, \ldots u_n} \int_{H_u} \left(\phi^+\left(x-\frac{u}{a_{n,d}}\right)-1\right)dx &\geq (a_{n,d}^d-n) (4a_{n,d})^{-s} (2a_{n,d})^{-d}\\
    &\geq 2^{-3s-2d}5^{-s/d}n^{-s/d},
\end{align*}

\noindent
where the second inequality used the fact: $a_{n,d} \leq (5n)^{1/d}\times 2$.



\noindent
Hence, there exists $N(s,d)$, such that for $n$ larger than $N(s,d)$,
\begin{align*}
\E_{f  \sim \mathcal{D}_{\mathcal{F}_1}}(L_n'(\mathfrak{s}'))  \geq 3^{-1}2^{-1-3s-2d}5^{-s/d}n^{1-s/d}.
\end{align*}



\endproof
\clearpage
\newpage

\end{document}